\documentclass[10pt,twocolumn,letterpaper]{article}

\usepackage{iccv}
\usepackage{times}
\usepackage{epsfig}
\usepackage{graphicx}
\usepackage{amsmath}
\usepackage{amssymb}
\usepackage{chen}

\newcommand{\pub}[1]{{\color{gray}{\tiny{[{#1}]}}}}
\def\Ours{{\textsc{LogicDiag}}}

\usepackage{algorithm}
\usepackage[accsupp]{axessibility}
\usepackage{etoolbox}
\makeatletter
\AfterEndEnvironment{algorithm}{\let\@algcomment\relax}
\AtEndEnvironment{algorithm}{\kern2pt\hrule\relax\vskip3pt\@algcomment}
\let\@algcomment\relax
\newcommand\algcomment[1]{\def\@algcomment{\footnotesize#1}}
\renewcommand\fs@ruled{\def\@fs@cfont{\bfseries}\let\@fs@capt\floatc@ruled
  \def\@fs@pre{\hrule height.8pt depth0pt \kern2pt}%
  \def\@fs@post{}%
  \def\@fs@mid{\kern2pt\hrule\kern2pt}%
  \let\@fs@iftopcapt\iftrue}
\makeatother

\usepackage[british, english, american]{babel}
\usepackage[pagebackref=true,breaklinks=true,letterpaper=true,colorlinks,urlcolor=gray,bookmarks=false]{hyperref}
\iccvfinalcopy

\def\iccvPaperID{}

\ificcvfinal\fi

\begin{document}

\title{Logic-induced Diagnostic Reasoning for Semi-supervised Semantic Segmentation}

\author{%
Chen Liang, \quad Wenguan Wang,  \quad Jiaxu Miao, \quad Yi Yang\thanks{Corresponding author: Yi Yang.} \vspace{.5em} \\
  \small{ReLER, CCAI, Zhejiang University} \vspace{.5em} \\
  \small\url{https://github.com/leonnnop/LogicDiag} \vspace{.5em}
}

\maketitle
\ificcvfinal\thispagestyle{empty}\fi

\begin{abstract}
   Recent advances in semi-supervised semantic segmentation have been heavily reliant on pseudo labeling to compensate for limited labeled data, 
   disregarding the valuable relational knowledge among semantic concepts.
   To bridge this gap, we devise {\Ours}, a brand new neural-logic semi-supervised learning framework. Our key insight is that conflicts within pseudo labels, identified through symbolic knowledge, can serve as strong yet commonly ignored learning signals.
   {\Ours} resolves such conflicts via reasoning with logic-induced diagnoses,
   enabling the recovery of (potentially) erroneous pseudo labels, ultimately alleviating the notorious error accumulation problem.
   We showcase the practical application of {\Ours} in the data-hungry segmentation scenario, where we formalize the structured abstraction of semantic concepts as a set of logic rules.
   Extensive experiments on three standard semi-supervised semantic segmentation benchmarks demonstrate the effectiveness and generality of {\Ours}.
   Moreover, {\Ours} highlights the promising opportunities arising from the systematic integration of symbolic reasoning into the prevalent statistical, neural learning approaches.
\end{abstract}

\vspace{.1em}
\section{Introduction}
\label{sec:intro}

\begin{figure}[t]
   \begin{center}
       \includegraphics[width=\linewidth]{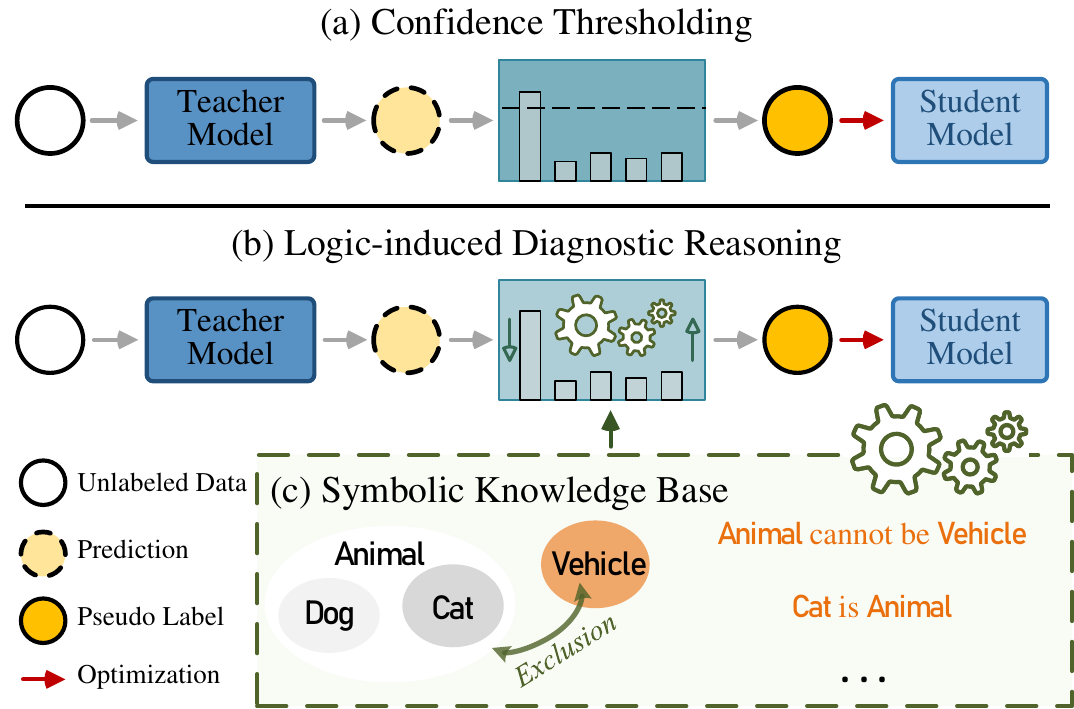}
        \put(-88, 28){\fontsize{5.5pt}{1em}\selectfont$\forall x\,\texttt{Animal}(x)\!\to\!\neg\texttt{Vehicle}(x)$}
        \put(-78, 12.5){\fontsize{5.5pt}{1em}\selectfont$\forall x\,\texttt{Cat}(x)\!\to\!\texttt{Animal}(x)$}
   \end{center}
   \vspace{-11pt}
   \captionsetup{font=small}
   \caption{\small Prevalent data-driven pseudo-labeling methods typically rely on heuristic confidence thresholding (\ie, (a)). We opt to le- verage symbolic knowledge in the form of logic rules (\ie, (c)), to diagnose and resolve potential errors within predictions (\ie, (b)).}
   \vspace{-3pt}
   \label{fig:motivation}
\end{figure}

Deep learning has revolutionized computer vision tasks, yielding remarkable breakthroughs~\cite{he2016deep,long2015fully,girshick2015fast,liu2021swin,yang2021multiple}. However, such advances are often only possible in the presence of large labeled training datasets, which are challenging to acquire.
Semantic segmentation, in particular, poses difficulties due to the need for pixel-level manual labeling~\cite{bearman2016s,wang2021exploring,lianggmmseg,liang2023clustseg,zhou2022rethinking,miao2021vspw,wang2022looking,liang2023local}, which is time-consuming and labor-intensive, precluding the application of such methods, especially in domains like medical image analysis.
{To remedy this issue}, there has been a growing interest in semi-supervised semantic segmentation, which aims to train segmentation models using a combination of limited labeled data and a large amount of unlabeled data~\cite{yuan2022semi,hu2021semi,jin2022semi,tarvainen2017mean,yang2021collaborative,yang2021associating,yang2022decoupling}.
Pseudo labeling~\cite{lee2013pseudo,sohn2020fixmatch} constitutes one such technique, where the unlabeled data is assigned pseudo labels based on model predictions. The model is then iteratively trained using these pseudo labeled data as if they were labeled examples. Typically, this method is implemented within the teacher-student framework~\cite{tarvainen2017mean} (Fig.\,\ref{fig:motivation}\,(a)).

Despite its prevalence, the pseudo labeling paradigm, a statistical learning approach, is commonly perceived as unreliable, {due to} the accumulation of erroneous predictions~\cite{arazo2020pseudo}. Previous works~\cite{sohn2020fixmatch,berthelot2019mixmatch,zhang2021flexmatch} have attempted to mitigate this issue by rejecting pseudo labels with classification scores below a heuristic threshold, known as confidence thresholding~\cite{sohn2020fixmatch}.
However, relying solely on this strategy often proves unsatisfactory.
\textbf{First}, the confirmation bias that may occur during the early stages of training can be difficult, if not impossible, to be rectified in subsequent learning~\cite{chen2022debiased}.
\textbf{Second}, the purely data-driven nature of such threshold-based methods makes them challenging to interpret.
\textbf{Third}, to achieve optimal performance, the threshold must be manually adjusted for each model and dataset, lim- 
\newpage
\noindent iting its practical application.
Beyond the confines, symbolic reasoning, as a compelling alternative, offers appealing characteristics: it requires \textit{little} or \textit{no data} to generalize systematically.
In light of this background, we suggest a promising direction towards an integration of statistical learning and symbolic reasoning to harness the strengths of both paradigms to achieve improved  performance~\cite{fang2018learning,wang2018attentive}.

In this paper, we propose to explicitly compile the rich symbolic knowledge into the prevalent 
pseudo label based neural training regime (\textit{cf.\!}~Fig.\,\ref{fig:motivation}\,(b)).
Our key insight is that symbolic knowledge can effectively resolve conflicts within the pseudo labels, which reveal potential model errors.
For instance, leveraging the prior knowledge on compositionality, we can naturally identify the inconsistencies like classifying a pixel as both a \texttt{cat} and a \texttt{vehicle} (\textit{cf.\!}~Fig.\,\ref{fig:motivation}\,(c)).
Correcting these errors progressively enhances the accuracy of pseudo labels, and thus mitigating the confirmation bias in iterative self-training.
Building upon this insight, we introduce {\Ours}, a new SSL framework that employs logical reasoning to identify such conflicts based on the symbolic knowledge expressed in the form of first-order logic. The framework then suggests possible diagnoses to rectify the prediction. 
To further address the challenge of multiple diagnoses, we model the likelihood of each diagnosis being the actual fault based on a comprehensive measure of both predictive confidence and degree of conflicts according to the fuzzy logic.
By doing so, our approach introduces a holistic neural-logic machine, that consolidates the benefits of powerful declarative languages, transparent internal functionality, and enhanced model performance.

{\Ours} is a principled framework that seamlessly integrates with mainstream semi-supervised learning methods. It requires only minor adjustments to the dense classification head. 
With {\Ours}, we can easily describe diverse symbolic knowledge using first-order logic and inject it into sub-symbolic pipelines.
Taking dense segmentation as main battlefield, we capture and evaluate a central cognitive ability of human, \ie, structured abstractions of visual concepts~\cite{bill2020hierarchical}, by grounding three logic rules onto {\Ours}:
\textit{Composition}, \textit{Decomposition}, and \textit{Exclusion}.

Extensive experiments have validated the effectiveness of our {\Ours}, which exhibit solid performance gains (\ie, 1.21\%-4.15\% mIoU) on three well-established benchmarks (\ie, PASCAL VOC 2012~\cite{everingham2015pascal}, Cityscapes~\cite{cordts2016cityscapes}, and COCO~\cite{lin2014microsoft}). We particularly observe significant improvements in the label scarce settings, indicating {\Ours}'s superior utilization of unlabeled data.
Besides, when employed onto the existing SSL frameworks, \eg, AEL~\cite{hu2021semi}, MKD~\cite{yuan2022semi}, the performance is consistently advanced.
The results demonstrate the strong generality and promising performance of {\Ours}, that also evidence the great potential of the integrated neural-symbolic computing in the fundamental large-scale semi-supervised segmentation.

\vspace{-5pt}
\section{Related Work}\label{sec:related}
\vspace{-3pt}

\noindent\textbf{Semi-Supervised Semantic Segmentation.}
Recent years have witnessed remarkable progress in image-level semi-supervised learning, driven primarily by two paradigms: self-training~\cite{nigam2000analyzing,grandvalet2005semi,lee2013pseudo} and consistency regularization~\cite{bachman2014learning,sajjadi2016regularization}.
Despite the impressive outcomes they produce, both paradigms heavily rely on the quality of pseudo-labels generated by the network itself, which makes them susceptible to confirmation bias~\cite{arazo2020pseudo} and leads to error accumulation throughout training.
Considerable efforts have been devoted to addressing this issue through confidence thresholding~\cite{berthelot2019remixmatch,sohn2020fixmatch,xie2020unsupervised,xie2020self}, curriculum scheduling~\cite{zhang2021flexmatch,xu2021dash}, sample ensembling across training iterations~\cite{laine2016temporal,tarvainen2017mean}, multiple augmented views~\cite{berthelot2019mixmatch}, and/or neighboring examples~\cite{iscen2019label,shi2018transductive,li2021comatch}. Some others introduce perturbations in data~\cite{xie2020unsupervised,sohn2020fixmatch,xie2020self}, feature~\cite{kuo2020featmatch}, and/or network~\cite{rasmus2015semi,sajjadi2016regularization}.
Similarly, in the context of semi-supervised semantic segmentation, which provides a pixel-level interpretation of visual semantics, prevalent approaches also follow the self-training paradigm~\cite{feng2022semi,yuan2021simple,yang2021st++,zhai2019s4l} and consistency regularization paradigm~\cite{french2020semi,ke2020guided,zou2020pseudoseg,ouali2020semi,kim2020structured,lai2021semi}. Very recent endeavors have further pushed forward the frontier through incorporating dense perturbations~\cite{yuan2022semi,yang2021st++,tu2022guidedmix,liu2022perturbed,zou2020pseudoseg}, contrastive supervisions~\cite{wang2022semi,liu2021bootstrapping,zhong2021pixel,zhou2021c3}, or coping with imbalanced and long-tailed nature of pixel samples~\cite{fan2022ucc,hu2021semi}.

Although significant advancements have been made, the problem of error accumulation remains far from being fully resolved. Existing methods heavily rely on an empirical threshold to justify pseudo labels, lacking explicit manipulation of rich symbolic knowledge. One common oversight in existing methods is the neglect of informative structures between semantic concepts.
As a consequence, these approaches often yield barely satisfactory results in terms of both model performance and generality.
One notable exception is~\cite{garg2022hiermatch}. While a label hierarchy is incorporated, it is only used for reducing labeling costs of fine-grained classes, resulting in limited improvement.
This work pursues an integrated neural-logic framework that addresses these fundamental limitations, providing a refreshing viewpoint on semi-supervised semantic segmentation.

\noindent\textbf{Hierarchical $_{\!}$Classification.} $_{\!}$Class-wise $_{\!}$hierarchical $_{\!}$depen- dencies have been studied in supervised tasks across several machine $_{\!}$learning $_{\!}$domains, $_{\!}$\eg, $_{\!}$functional genomics\tcite{barutcuoglu2006hierarchical,cerri2012genetic,li2018deepre}, text categorization~\cite{rousu2006kernel}, object recognition~\cite{marszalek2007semantic,grauman2011learning,hwang2012semantic}, image classification~\cite{bengio2010label,deng2010does}.
In the computer vision field, the class taxonomy is mainly explored through: i) semantic-aligned \textit{label embedding}~\cite{xian2016latent,frome2013devise,akata2015evaluation}; ii) hierarchy-coherent \textit{loss constraint}~\cite{deng2010does,verma2012learning,zhao2011large}; or iii) structured \textit{network architecture}~\cite{ahmed2016network,yan2015hd,zweig2007exploiting}.
Among the previous efforts, only a few attempts~\cite{li2022deep,wang2020hierarchical,wang2021hierarchical,wang2019learning,saha2022improving} towards label hierarchy-aware semantic segmentation have been made. Whereas, they all heavily rely on labeled hierarchical data and/or specialized neural architectural design, making them impractical for use when merely a handful of labeled data is available, as in SSL.
In contrast, our algorithm exploits class taxonomy as the symbolic knowledge for diagnosing conflicts in model predictions, enabling comprehensive exploration of hierarchical relations in both labeled and unlabeled data. The general design also makes our algorithm versatile and easily implementable to standard hierarchy-agnostic SSL architectures.

\noindent\textbf{Neural-Symbolic Computing.}
Building preferable computational methods for integrated statistical learning and symbolic reasoning is a long-standing challenge~\cite{mcculloch1943logical}.
This active line of research, namely neural-symbolic computing (NSC),
draws soaring attention in recent years~\cite{giunchiglia2021multi,teru2020inductive,lin2019kagnet,rocktaschel2017end}. NSC shows great potential to reconcile the robust learning capabilities of neural networks with the interpretability and reasoning abilities of symbolic representations~\cite{garcez2022neural,wang2022towards}, and thereby gains widespread recognition as a catalyst for the next generation of AI~\cite{lake2017building,marcus2018deep}.
NSC has demonstrated its superiority across a wide range of domains, including mathematical reasoning~\cite{lample2019deep,li2020closed,arabshahi2018combining}, robotics control~\cite{sun2021neuro,silver2022learning,zhu2021hierarchical}, as well as scientific discovery~\cite{cranmer2020discovering,jumper2021highly,shah2020learning,tseng2022automatic}.
Its virtue of data efficiency has also attracted researchers from SSL fields~\cite{marra2020integrating,van2019semi,xu2018semantic,tsamoura2021neural}, where the symbolic knowledge, usually expressed in logic, is mostly incorporated as a form of regularization through loss constraints applied to output space.
Though being exciting, the advances have been primarily limited to ``toy'' tasks.
The full potential and challenges of NSC for large-scale realistic problems remain largely unexplored.
Our method is partly motivated by, but also distinct from, previous efforts that merely encourage valid output structures in a soft manner, thereby allowing errors to accumulate, especially during the initial stages of training.

To $_{\!}$our $_{\!}$best $_{\!}$knowledge, $_{\!}$this $_{\!}$is $_{\!}$the $_{\!}$first $_{\!}$work $_{\!}$that $_{\!}$promotes $_{\!}$and $_{\!}$implements $_{\!}$an $_{\!}$integrated $_{\!}$neural-symbolic $_{\!}$framework $_{\!}$in\\ large-scale vision-oriented SSL.
Previous studies have primarily concentrated on the neural aspect, \ie, sub-symbolic methods.
In contrast, our method explicitly compiles symbolic knowledge into training regime of the neural network. This integration allows us to leverage the advantages of powerful declarative languages and transparent internal functionality.
The encouraging results we have obtained provide compelling empirical evidence of the significant potential of NSC in the large-scale vision domain.

\vfill
\section{Methodology}\label{sec:method}
We commence by formalizing modern semi-supervised semantic segmentation approaches, situating them within a sub-symbolic framework and highlighting their inherent limitations (\S\ref{sec:meth:ssl}).
We then present {\Ours}, a general logic-induced diagnosis framework, that complements the current sub-symbolic approaches with a principled infusion of symbolic knowledge in the form of logic rules (\S\ref{sec:meth:define}).
Finally, we showcase its practical application in the realm of visual semantic interpretation (\S\ref{sec:meth:use}).

\pagebreak 
\noindent\textbf{Problem Statement.} In the standard SSL setting, given an unknown distribution over visual space $\Xcal$ (\eg, pixel space for segmentation) and category label space $\Ycal\!=\!\{1,\cdots\!,C\}$ with $C$ semantic categories, the goal is to find a predictor $h\!:\Xcal\!\mapsto\!\Ycal$, such that the generalization error is minimized, based on the observed data $\Dcal$, consisting of a labeled subset $\Dcal^{l}\!=\!\{(x_{i}, y_i)\!\in\!\Xcal\!\times\!\Ycal\}_{i=1}^{N^l}$ and an unlabeled subset $\Dcal^{u}\!=\!\{u_{j}\!\in\!\Xcal\}_{j=1}^{N^u}$. Typically, $N^l$ is rather small, \ie, $N^u\!\gg\!N^l$, resulting in the issue of insufficient learning signal.

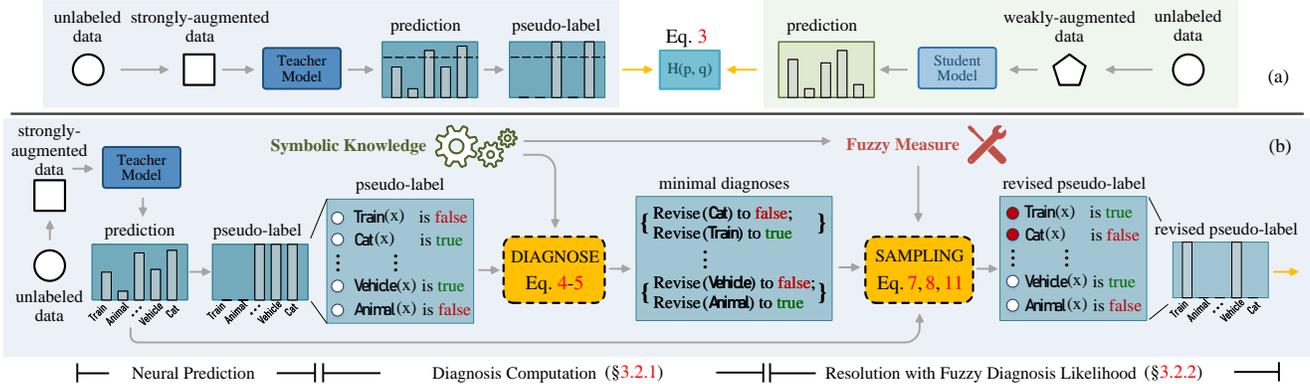
\begin{figure*}[t!]
   \begin{center}
      \includegraphics[width=1.\linewidth]{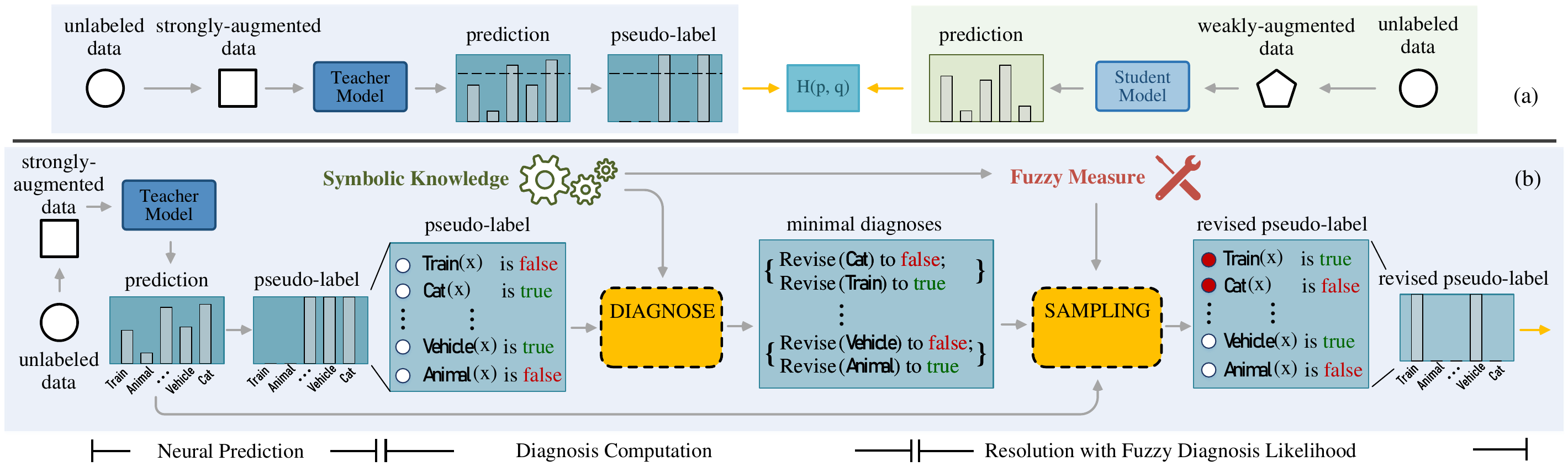}
      \put(-245, 130){\fontsize{7.5pt}{1em}\selectfont Eq.~\ref{eq:floss}}
      \put(-268, 3){\scriptsize (\S\ref{sec:meth:define:DC})}
      \put(-64, 3){\scriptsize (\S\ref{sec:meth:measure})}
      \put(-298, 36.5){\fontsize{7.5pt}{1em}\selectfont Eq.~\ref{eq:conflict}-\ref{eq:diagnosis}}
      \put(-165, 36.5){\fontsize{7.5pt}{1em}\selectfont Eq.\,\ref{eq:likelihood},\,\ref{eq:fail},\,\ref{eq:truth_degree}}
   \end{center}
   \vspace{-14pt}
   \captionsetup{font=small}
   \caption{\small Illustrations of (a) sub-symbolic SSL pipeline (\S\ref{sec:meth:ssl}); (b) {\Ours}, logic-induced diagnostic reasoning framework (\S\ref{sec:meth:define}).}
   \label{fig:pipeline}
   \vspace{-4pt}
\end{figure*}

\subsection{Sub-symbolic Semi-supervised Learning}\label{sec:meth:ssl}
\noindent Mainstream solutions predominantly revolve around a neural pipeline, known as sub-symbolic methods~\cite{garcez2019neural}. 
They rely on the consistency regularization paradigm~\cite{sohn2020fixmatch,chen2021semi}, that comprises four key components, as shown in Fig.\,\ref{fig:pipeline}\,(a).
\begin{itemize}[leftmargin=*]
   \setlength{\itemsep}{0pt}
   \setlength{\parsep}{-2pt}
   \setlength{\parskip}{-0pt}
   \setlength{\leftmargin}{-10pt}
   \vspace{-6pt}%

   \item \textit{Data augmentors} $\Acal(\cdot)$/$\alpha(\cdot)$, that transform examples into strongly-/weakly-augmented views, respectively.
   \item \textit{Base encoder} $f(\cdot)$, that maps augmented views into $D$-dimensional representations, \ie, $\bm{x}\!=\!f\circ\alpha(x)\!\in\!\RR^D$. 
   \item \textit{Prediction head} $g(\cdot)$, that gives the predictive probability distribution from the representation, \ie, $\obm\!=\!g(\bm{x})\!\in\!\Delta^C$, and $\Delta^C$ is the $C$-way probability simplex.
   \item \textit{Pseudo label processor} $\psi(\cdot)$, that converts the raw predictions into pseudo labels guided by heuristic priors or assumptions, \eg, confidence thresholding~\cite{sohn2020fixmatch,berthelot2019mixmatch,berthelot2019remixmatch,zhang2021flexmatch}.

   \vspace{-6pt}
\end{itemize}
The predictor $h$, consisting of the base encoder and prediction head (\ie, $h\!=\!g\circ f$), is jointly optimized on the complete observed set $\Dcal$, \ie, $\{\Dcal^l, \Dcal^u\}$.
For the labeled ones, the standard cross entropy, {denoted by $\text{H}(\cdot,\cdot)$}, can be directly applied on the weakly augmented examples:
\vspace{-3pt}
\begin{equation}\small\label{eq:fsup}
\begin{aligned}\small
    \Lcal^l=\dfrac{1}{N^l} \sum\nolimits_{i=1}^{N^l}\text{H}\big((h \circ \alpha)(x_i), {y}_i\big).
\end{aligned}
\vspace{-3pt}
\end{equation}
For the unlabeled data, the pseudo labels are generated from the weakly augmented views of given examples. As a prevalent choice~\cite{chen2021semi}, the pseudo label processor $\psi(\cdot)$ filters out unreliable pseudo labels with a confidence threshold $\tau$, \ie, $\psi(\obm)\!=\!\mathbb{I}(\max(\obm)\!\geq\!\tau)\!\cdot\!\argmax\obm$. Here $\mathbb{I}$ defines the identify function. Then, the predictor $h$ is expected to yield consistent predictions on the strongly augmented views:
\vspace{-3pt}
\begin{equation}\small\label{eq:funsup}
\begin{aligned}\small
   \Lcal^u = \dfrac{1}{N^u} \sum\nolimits_{j=1}^{N^u}\text{H}\big((h\circ\!\,\Acal)(u_j), (\psi\circ h\circ\alpha)(u_j)\big),
\end{aligned}
\vspace{-3pt}
\end{equation}
The entire network is supervised with both the two losses:
\vspace{-3pt}
\begin{equation}\small\label{eq:floss}
\begin{aligned}\small
   \Lcal = \Lcal^l + \lambda\Lcal^u,
\end{aligned}
\vspace{-3pt}
\end{equation}
where the scalar hyperparameter $\lambda$ trade off the two terms.
Despite the prevalence it achieved, there remains three issues in mainstream SSL methods:
\textbf{First}, they simply treat all categories equally in a flat view, overlooking the structured relationships between the visual concepts. The transferable knowledge residing in the label hierarchy is also discarded, which is of particular essential in the label scarce scenarios.
\textbf{Second}, self-training with pseudo labels in turn (Eq.~\ref{eq:funsup}) inevitably accumulates errors and causes confirmation bias~\cite{arazo2020pseudo}, which severely affects the model performance.
Although low-confidence filtering (\ie, $\psi$) can alleviate the problem, errors accumulated in the early training stage are difficult to be corrected in subsequent training, especially for the poorly-behaved categories~\cite{chen2022debiased}.
\textbf{Third}, the label filtering process relies on the empirical adjustment of a threshold $\tau$, which requires specialization for each model and dataset to promote the performance. This significantly limits the flexibility and generality of current methods.

Accordingly, we suggest that it is now imperative to rethink prevailing sub-symbolic pipeline, in which the filtering process $\psi(\cdot)$ has been more detrimental than beneficial.

\vspace{6pt}
\subsection{Logic-induced Diagnostic Reasoning}\label{sec:meth:define}
\noindent A well-performed SSL model not only generates accurate sample labels but also aligns with our background knowledge of the world, \eg, \textit{bird is an animal}, \textit{sky is above the grass}.
By resolving conflicting predictions through reasoning based on the knowledge, we can potentially correct erroneous pseudo-labels, which leads to a more effective learning process of the neural aspect (\ie, $h(\cdot)$) and ultimately reduces the accumulation of errors.
To this end, our approach, {\Ours}, incorporates the symbolic aspect (\ie, $\psi(\cdot)$), which reasons about malfunctions, dubbed \textit{conflict}, in the model's output (\ie, pseudo label) according to the symbolic knowledge, and offers possible solutions, dubbed \textit{diagnosis}~\cite{reiter1987theory}, to resolve the conflicts within predictions. 

\vspace{-3pt}
\subsubsection{Diagnosis Computation}\label{sec:meth:define:DC}
We formalize our target with a triple $\Scal\!=\!\left\langle \Dcal_x,h,\Kcal \right\rangle$:
\setlist{nosep}
\begin{itemize}[leftmargin=*]
   \setlength{\itemsep}{0pt}
   \setlength{\parsep}{-2pt}
   \setlength{\parskip}{-0pt}
   \setlength{\leftmargin}{-10pt}

   \item $\Dcal_x\!=\!\{x_i\}_{i=1}^{N^l\!+\!N^u}$ defines the collection of pixel samples from both labeled and unlabeled datasets;
   \item $h$ is the neural predictor, that generates a set of binary pseudo labels, denoted as $\Ocal\!=\!\{o_j\}_{j=1}^{|\Ocal|}$, with one label for each semantic concept such as \texttt{bird}, \texttt{animal}, \etc.
   Each $o$ determines whether a given pixel $x$ belongs to a specific concept or not, 
   which can be achieved by applying a binarizing operation $b$, to the output of $h$, \ie, $\Ocal\!=\!b\circ h(x)$.\footnote{Here we extend $h$ to generate binary outputs by attaching a sigmoid function, without modifying the main architecture of the network (\textit{cf.\!}~\S\ref{sec:meth:use}).}
   \item $\Kcal$ is a finite set of rules expressed in \textit{first-order logic} (FOL) that captures the world knowledge on normality.
\end{itemize}
We provide a gentle introduction on FOL, which comprises four parts:
\textbf{i)} \textit{constants} representing specific pixel instances $x_i$ or truth (\ie, $\top$: true, $\bot$: false);
\textbf{ii)} \textit{variables} ranging over these constants, denoted by $x$;
\textbf{iii)} \textit{predicates} evaluating the semantics of variables to be true or false (\eg, $\texttt{bird}(x)$ is true states the fact that pixel $x$ belongs to a category \texttt{bird});
\textbf{iv)} \textit{connectives} (\eg, $\wedge$: and, $\vee$: or, $\neg$: not, $\to$: imply) and \textit{quantifiers} (\ie, $\forall$: for all, $\exists$: exist) over finite predicates.
To simplify subsequent formulations without altering the meaning, we may omit the explicit mention of the pixel variable $x$ in certain predicates (\eg, $o(x)$ to $o$).

A \textit{conflict} arises when the assumption that all outputs are normal is inconsistent with our symbolic knowledge $\Kcal$:
\vspace{-3pt}
\begin{equation}\small\label{eq:conflict}
\begin{aligned}\small
   \Kcal \wedge \Ocal \wedge \bigwedge\nolimits_{o\in \Ocal}\!t(o) \vdash \bot.
\end{aligned}
\vspace{-3pt}
\end{equation}
Here $\vdash$ is logical entailment. We define the unary predicate $t(\cdot)$ over the output $o$ such that $t(o)$ is true when $o$ is normal in terms of consistency, while $\neg t(o)$ is true when $o$ is faulty.
To explain the inconsistency above, a \textit{diagnosis} is proposed by assuming that some outputs in a set $\omega\!\subseteq\!\Ocal$ are faulty:
\vspace{-3pt}
\begin{equation}\small\label{eq:diagnosis}
\begin{aligned}\small
   \Kcal \wedge \Ocal \wedge \bigwedge\nolimits_{o\in \omega}\!\!\!\neg\,t(o) \wedge \bigwedge\nolimits_{o\in \Ocal \setminus \omega}\!t(o) \nvdash \bot.
\end{aligned}
\vspace{-3pt}
\end{equation}
For all possible diagnoses, we restrict our focus solely to the ones that contain only false elements, in the world being modeled, known as minimal diagnosis~\cite{reiter1987theory}. Formally, a diagnosis $\omega$ is minimal \textit{iff.} any strict subset $\omega'\!\subset\!\omega$ is not a diagnosis.
Thus far, the problem has been reduced to a Boolean satisfiability problem. We present to solve it with greedy algorithms.
In practice, considering the rather small search space (\ie, $|\Kcal|$ and $|\Ocal|$), we keep our pipeline straightforward without employing sequential approximation (\eg, MCMC).
At this point, a diagnosis has already been computed and the pseudo label $\Ocal$ can be revised to the logically consistent $\Ocal'$ according to the following equation:
\vspace{-3pt}
\begin{equation}\small\label{eq:resolution}
\begin{aligned}\small
   \Ocal'\!=\!\{\neg o\}_{o\in\omega}\cup\{o\}_{o\in \Ocal\setminus\omega}.
\end{aligned}
\end{equation}

However, there might be multiple diagnoses that are consistent with our target,
it is not immediately clear which one is the `correct' revision to make.
A na\"ive approach would be to uniformly sample from all possible minimal diagnoses and revise them one at a time, which is inefficient and ineffective, as demonstrated by the empirical results (\textit{c.f.\!}~\S\ref{sec:exp:dig}).
To address this issue, we take one step further to model the likelihood of the diagnosis being the actual faulty.

\subsubsection{Resolution with Fuzzy Diagnosis Likelihood}\label{sec:meth:measure}

We can derive the likelihood of a correct diagnosis given the triple system $\Scal\!=\!\left\langle \Dcal_x,h,\Kcal \right\rangle$ under the common assumption of independent failure~\cite{rodler2022random}, using the multiplication rule:
\vspace{-3pt}
\begin{equation}\small\label{eq:likelihood}
\begin{aligned}\small
   P(\omega|\Scal) = \prod\nolimits_{o\in\omega^{\!+}}\!P(t(o)|\Scal) \,\prod\nolimits_{o\in\omega^{\!-}}\!\big(1\!-\!P(t(o)|\Scal)\big),
\end{aligned}
\vspace{-3pt}
\end{equation}
where $\omega$ is the diagnosis; $\omega^+$ and $\omega^-$ are the sets of outputs assigned normal and abnormal behavior modes, respectively.
To comprehensively estimate the probability of an output $o$ being normal, denoted as $P(t(o)|\Scal)$, we consider both the degree of conflict and predictive confidence:
\vspace{-3pt}
\begin{equation}\small\label{eq:fail}
\begin{aligned}\small
   P(t(o)|\Scal) \coloneqq \!\left\{\begin{matrix}
      P(o|\Scal) \cdot \big(1-c(o\,;\Scal)\big),~~~~~~~~~~~~\text{if}~o \vdash \top \\
      \big(1-P(o|\Scal)\big) \cdot \big(1-c(o\,;\Scal)\big),~\text{if}~o \vdash \bot
    \end{matrix}\right.
\end{aligned}
\end{equation}
Here, function $c$ maps each output $o\!\in\!\Ocal$ to its conflict degree $c(o\,;\Scal)\!\in\![0,1]$; and $P(o|\Scal)\!=\!P(o|x)\!=\!h(x)\!\in\![0,1]$ represents the predictive probability.
Intuitively, if an output violates the established knowledge but has a high predictive confidence, it should be penalized more than the others (\ie, a small valued $c(o\,;\Scal)$), increasing the likelihood of correct diagnosis.
Conversely, high confidence indicates low probability in abnormality, thereby balancing the likelihood.

To measure this degree of conflict, we resort to the fuzzy logic~\cite{kosko1993fuzzy}, a form of soft probabilistic logic, specifically, the \textit{Goguen fuzzy logic}~\cite{hajek2013metamathematics} and \textit{G\"odel fuzzy logic}~\cite{godel1986kurt}.
Fuzzy logic generalizes FOL to uncertain inputs, where the variables have truth values in $[0,1]$, \eg, the predictive probability with respect to each visual concept in our case.
The logical connectives (\eg, $\wedge,\vee,\neg$) are approximated with fuzzy operators (\ie,  \textit{t-norm}, \textit{t-conorm}, \textit{fuzzy negation}):
\vspace{-3pt}
\begin{equation}\small\label{eq:connective}
\begin{aligned}\small
   \phi \wedge \varphi \coloneqq \phi \cdot \varphi,~~~ \phi \vee \varphi \coloneqq \max(\phi,\varphi),~~~ \neg\phi \coloneqq 1 - \phi.
\end{aligned}
\vspace{-3pt}
\end{equation}
Besides, the existential and universal quantifier (\ie, $\forall,\exists$) are approximated in a form of generalized mean~\cite{van2022analyzing,badreddine2022logic}:
\vspace{-3pt}
\begin{equation}\small\label{eq:quantifier}
\begin{aligned}\small
   \exists\phi(x) &\coloneqq \big(\frac{1}{|\Dcal_x|}\sum\nolimits_{x\in\Dcal_x}\phi(x)^q\big)^\frac{1}{q}, \\
   \forall\phi(x) &\coloneqq 1-\big(\frac{1}{|\Dcal_x|}\sum\nolimits_{x\in\Dcal_x}(1-\phi(x)^q)\big)^\frac{1}{q},
\end{aligned}
\end{equation}
where $q\!\in\!\ZZ$. Instead of being strictly true or false, fuzzy logic offers a soft measure on how much a logic rule is fired, \ie, truth degree, which can be interpreted as the complement to the desired conflict degree:
\vspace{-3pt}
\begin{equation}\small\label{eq:truth_degree}
\begin{aligned}\small
   c(o\,;\Scal) = 1-\frac{1}{|\Kcal|}\sum\nolimits_{k\in\Kcal} g(o,k;\Scal),
\end{aligned}
\vspace{-3pt}
\end{equation}
where we define $g(\cdot)$ as the fuzzy truth measurement.

Following the derivation, we are now ready to assess the extent to which a series of outputs violate the rules in our knowledge $\Kcal$ (\textit{c.f.\!}~Eq.~\ref{eq:truth_degree}), so as to estimate the likelihood of actual diagnoses (\textit{c.f.\!}~Eq.~\ref{eq:likelihood}-\ref{eq:fail}).
Then, the na\"ive utilization is to select the diagnosis with the highest probability concerning pixel $x_i$, \ie, $\argmax_{\omega_i} P(\omega_i|x_i,h,\Kcal)$. However, such local estimation is prone to get stuck in spurious corrections, as evidenced by our experiments (\textit{c.f.\!}~\S\ref{sec:exp:dig}).
Instead, we resort to Monte Carlo estimation to calculate the posterior distribution. Our empirical findings suggest that optimizing with just one sample per datapoint is sufficient.

Overall, as shown in Fig.\,\ref{fig:pipeline}\,(b), the reasoning and learning aspects of {\Ours} work iteratively. The neural model $h(\cdot)$ first poses assumptions about current observations, then the symbolic model $\psi(\cdot)$ reasons over the symbolic knowledge $\Kcal$ to determine the diagnoses, which in turn facilitates the learning process of the neural model.

\begin{figure*}[t]
   \begin{center}
      \includegraphics[width=1.\linewidth]{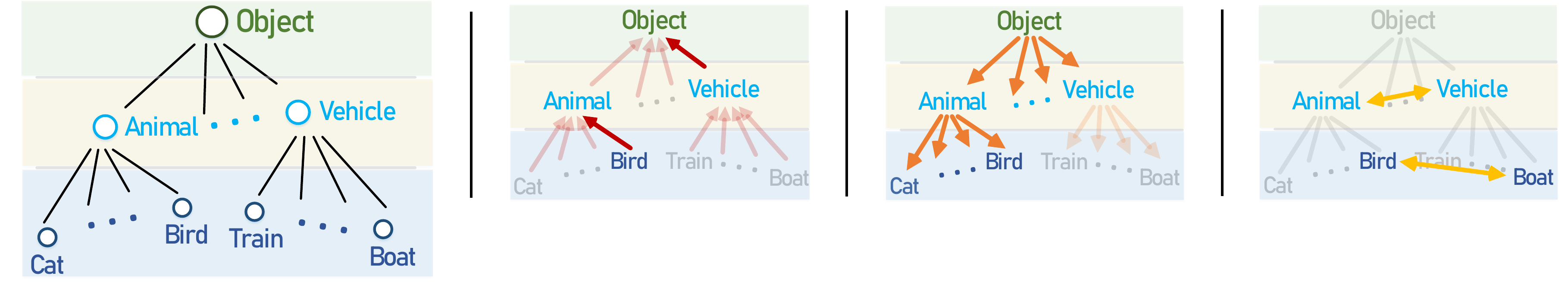}
      \put(-365, 77){\small $\Vcal_L$}
      \put(-365, 47){\small $\Vcal_l$}
      \put(-365, 17){\small $\Vcal_1$}
      \put(-337, 5){\fontsize{6.5pt}{1em}\selectfont \bf $\forall x\,\texttt{Bird}(x)\!\to\!\texttt{Animal}(x)$}
      \put(-337, 15){\fontsize{6.5pt}{1em}\selectfont \bf $\forall x\,\texttt{Vehicle}(x)\!\to\!\texttt{Object}(x)$}
      \put(-238, 5){\fontsize{6.5pt}{1em}\selectfont \bf $\forall x\,\texttt{Animal}(x)\!\!\to\!\!\texttt{Cat}(x)\!\vee\!\!...\!\!\vee\!\texttt{Bird}(x)$}
      \put(-238, 15){\fontsize{6.5pt}{1em}\selectfont \bf $\forall x\,\texttt{Object}(x)\!\!\to\!\!\texttt{Animal}(x)\!\vee\!\!...\!\!\vee\!\texttt{Vehicle}(x)$}
      \put(-95, 5){\fontsize{6.5pt}{1em}\selectfont \bf $\forall x\,\texttt{Bird}(x)\!\to\!\neg\,\texttt{Boat}(x)$}
      \put(-95, 15){\fontsize{6.5pt}{1em}\selectfont \bf $\forall x\,\texttt{Animal}(x)\!\to\!\neg\,\texttt{Vehicle}(x)$}
      \put(-490, 78){\small (a)}
      \put(-333, 78){\small (b)}
      \put(-215, 78){\small (c)}
      \put(-100, 78){\small (d)}
   \end{center}
   \vspace{-16pt}
   \captionsetup{font=small}
   \caption{\small Illustrations of (a) label hierarchy $\Tcal$; (b) Composition (Eq.~\ref{eq:rule:composition}); (c) Decomposition (Eq.~\ref{eq:rule:decomposition}); (d) Exclusion (Eq.~\ref{eq:rule:exclusion}) rules (\S\ref{sec:meth:use}).}
   \label{fig:rules}
\end{figure*}
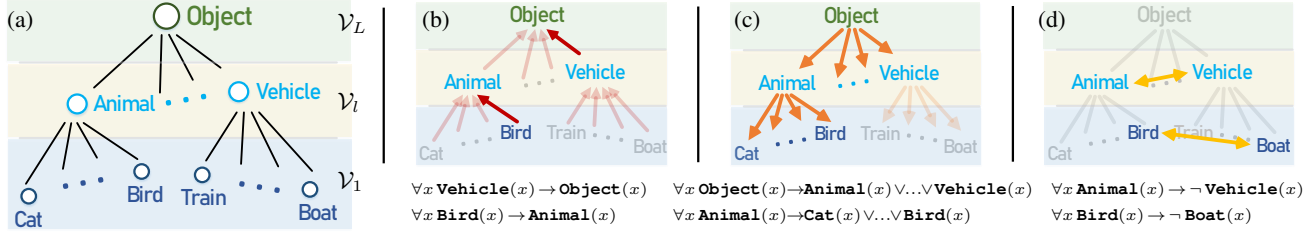

\subsection{Diagnosis with Visual Semantics: An Example}\label{sec:meth:use}
{\Ours} is model-agnostic to the neural aspect of SSL pipelines, and is also compatible with general symbolic knowledge described in FOL. As a result, {\Ours} can further be advanced by embracing the development of new architectures or incorporating more knowledge.
To showcase the practical deployment of {\Ours}, we examine its application in segmentation scenario, with particular interest in structured visual semantics~\cite{li2022deep} that are commonly overlooked in the mainstream. 
Specifically, the semantic concepts and their relations are formed as a tree-shaped label hierarchy $\Tcal\!=\!\left\langle \Ocal,\Ecal \right\rangle$ (\textit{c.f.\!}~Fig.\,\ref{fig:rules}\,(a)).
The node set $\Ocal$ is the union of nodes from $L$ levels of abstraction, denoted as $\Ocal\!=\!\cup_{l=1}^L\Ocal_l$. The leaf nodes, $\Ocal_1$, represent the most specific concepts, namely category labels, such as \texttt{bird}, \texttt{cat}, where $\Ocal_1\!=\!\Ycal$, $|\Ocal_1|\!=\!C$. The internal nodes represent higher-level concepts such as \texttt{vehicle}, \texttt{animal}, and the root nodes $\Ocal_L$ represent the most abstract concepts, such as \texttt{object}. 
Besides, the edge set $\Ecal$ encodes relational knowledge among all these concepts, with directed edges $u\!\rightarrow\!v\!\in\!\Ecal$ denoting a \textit{part-of} relation between two concepts $u,v\!\in\!\Ocal$ in adjacent levels, \eg, \texttt{animal}$\rightarrow$\texttt{bird}.

Recall that we define the $\left\langle \text{data},\text{model},\text{knowledge}\right\rangle$ triple $\left\langle \Dcal_x,h,\Kcal \right\rangle$ for {\Ours} (\textit{c.f.\!}~\S\ref{sec:meth:define:DC}).
In the context of structured semantic concept exploration, the model $h$ yields $|\Ocal|$ binary pseudo labels, denoting nodes within the label hierarchy.
Besides, $\Kcal_{\Tcal}$ contains FOL rules, describing the structured symbolic knowledge according to the label hierarchy $\Tcal$. Inspired by previous efforts~\cite{li2022deep,deng2014large,bi2011multi} in hierarchical classification, we define $\Kcal_{\Tcal}$ with three types of rules, \ie, \textit{composition}, \textit{decomposition}, and \textit{exclusion}.

\noindent$\bullet$ \textbf{Composition Rule} ($\Kcal_\text{C}$). \textit{If one class is labeled true, its parent (\ie, superclass) is labeled true} (Fig.\,\ref{fig:rules}\,(b)):
\vspace{-3pt}
\begin{equation}\small\label{eq:rule:composition}
\begin{aligned}\small
   \forall x(o(x)\to p_o(x)),
\end{aligned}
\vspace{-3pt}
\end{equation}
where $p_o$ is the parent node of $o$ in $\Tcal$, \ie, $p_o\!\to\!o\!\in\!\Ecal$.
For example, \textit{``bird is (a subclass of) animal''} shall be interpreted as: $\forall x(\texttt{bird}(x)\to \texttt{animal}(x))$.

\noindent$\bullet$ \textbf{Decomposition $_{\!}$Rule} $_{\!}$($\Kcal_\text{D}$). $_{\!}$\textit{If $_{\!}$one $_{\!}$class $_{\!}$is $_{\!}$labeled $_{\!}$true,} $_{\!}$\textit{at least one of its children (\ie, subclasses) is true} (Fig.\,\ref{fig:rules}\,(c)):
\vspace{-3pt}
\begin{equation}\small\label{eq:rule:decomposition}
\begin{aligned}\small
   \forall x(o(x)\to \bigvee\nolimits_{\!r_o\in R_o}\!\!r_o(x)),
\end{aligned}
\end{equation}
where $R_o$ is the set of children node(s) of $o$ in $\Tcal$, \ie, $o\!\to\!r_o\!\in\!\Ecal$.
For example, \textit{``animal subsumes (is the superclass of) bird, dog, $\cdots$, cat''} shall be interpreted as:\\ $\forall x(\texttt{animal}(x)\to \texttt{bird}(x)\vee\texttt{dog}(x)\vee\cdots\vee\texttt{cat}(x))$.

\noindent$\bullet$ \textbf{Exclusion Rule} ($\Kcal_\text{E}$). \textit{If one class is labeled true, all its sibling classes are labeled false} (Fig.\,\ref{fig:rules}\,(d)):
\vspace{-3pt}
\begin{equation}\small\label{eq:rule:exclusion}
\begin{aligned}\small
   \forall x(o(x)\to \bigwedge\nolimits_{s_o\in S_o}\!\!\!\neg\,s_o(x)),
\end{aligned}
\vspace{-3pt}
\end{equation}
where $S_o$ is the set of sibling node(s) of $o$ in $\Tcal$.
For example, \textit{``bird cannot be train, $\cdots$, nor cat''} shall be interpreted as: $\forall x(\texttt{bird}(x)\to \neg\,\texttt{train}(x)\wedge\cdots\wedge\neg\,\texttt{cat}(x))$.

According to the fuzzy measure defined in Eq.~\ref{eq:connective}-\ref{eq:quantifier}, we derive the truth degrees as follows (Note, $\phi\!\to\!\varphi\!\Leftrightarrow\!\neg\phi\vee\varphi$):

\noindent$\bullet$ Composition Rule (Eq.~\ref{eq:rule:composition}): $g(o,\Kcal_\text{C}|\Scal)=$
\vspace{-3pt}
\begin{equation}\small\label{eq:fuzzy:composition}
\begin{aligned}\small
   1\!-\!\sbr{\frac{1}{|\Dcal_x|}\!\sum_{x\in\Dcal_x}\!\!(P(o|x)\!-\!P(o|x)\!\cdot\!P(p_o|x))^q}^{\frac{1}{q}}\!\!.
\end{aligned}
\vspace{-3pt}
\end{equation}

\noindent$\bullet$ Decomposition Rule (Eq.~\ref{eq:rule:decomposition}): $g(o,\Kcal_\text{D}|\Scal)=$
\vspace{-3pt}
\begin{equation}\small\label{eq:fuzzy:decomposition}
\begin{aligned}\small
   1\!-\!\sbr{\frac{1}{|\Dcal_x|}\!\sum_{x\in\Dcal_x}\!\!(P(o|x)\!-\!P(o|x)\!\cdot\!\max_{r_o\in R_o}\!P(r_o|x))^q}^{\frac{1}{q}}\!\!.
\end{aligned}
\vspace{-3pt}
\end{equation}

\noindent$\bullet$ Exclusion Rule (Eq.~\ref{eq:rule:exclusion}): $g(o,\Kcal_\text{E}|\Scal)=$
\vspace{-3pt}
\begin{equation}\small\label{eq:fuzzy:exclusion}
\begin{aligned}\small
   1\!-\!\frac{1}{|S_o|}\sum_{s_o\in S_o}\sbr{\frac{1}{|\Dcal_x|}\!\sum_{x\in\Dcal_x}\!\!(P(o|x)\!\cdot\!P(s_o|x))^q}^{\frac{1}{q}}\!\!,
\end{aligned}
\vspace{-3pt}
\end{equation}
where we translate \textit{one-vs-multiple} exclusion (Eq.~\ref{eq:rule:exclusion}) to the\\ equivalent \textit{one-vs-one} form to avoid sorites paradox~\cite{goguen1969logic}.

We are now prepared to proceed with the calculations (\textit{c.f.\!}~Eq.\,\ref{eq:likelihood},\,\ref{eq:fail},\,\ref{eq:truth_degree}) and sample from the diagnosis likelihood $P(\omega|\Scal)$, where the pseudo labels can be revised accordingly (\textit{c.f.\!}~Eq.\,\ref{eq:resolution}). Overall {\Ours} is supervised with Eq.\,\ref{eq:floss}.

\newcommand{\reshl}[2]{
{#1} \fontsize{7.5pt}{1em}\selectfont\color{mygreen}{$\!\uparrow\!$ {#2}}
}
\newcommand{\reshlb}[2]{
\textbf{#1} \fontsize{7.5pt}{1em}\selectfont\color{mygreen}{$\!\uparrow\!$ \textbf{#2}}
}
\begin{table*}[t]
   \centering\small
   \resizebox{1.\textwidth}{!}{
      \tablestyle{5.2pt}{1.05}
      \begin{tabular}{|rl||cccc|cccc|}
         \thickhline
         \rowcolor{mygray}
         & & \multicolumn{4}{c|}{PASCAL VOC 2012 \textit{original}} & \multicolumn{4}{c|}{PASCAL VOC 2012 \textit{augmented}} \\
         \rowcolor{mygray}
         \multicolumn{2}{|c||}{\multirow{-2}{*}{Method}} & 1/2 (732) & 1/4 (366) & 1/8 (183) & 1/16 (92) & 1/2 (5291) & 1/4 (2646) & 1/8 (1323) & 1/16 (662) \\

         \hline\hline
         MT~\cite{tarvainen2017mean}\!\!\!\!&\!\!\!~\pub{NeurIPS17}   & 69.16  & 63.01  & 55.81  & 48.70 & 77.61 & 76.62 & 73.20 & 70.59 \\
         GCT~\cite{ke2020guided}\!\!\!\!&\!\!\!~\pub{ECCV20}   & 70.67  & 64.71  & 54.98  & 46.04 & 77.14 & 75.25 & 73.30 & 69.77 \\
         CutMixSeg~\cite{french2020semi}\!\!\!\!&\!\!\!~\pub{BMVC20}   & 69.84  & 68.36  & 63.20  & 55.58 & 75.89 & 74.25 & 72.69 & 72.56 \\
         PseudoSeg~\cite{zou2020pseudoseg}\!\!\!\!&\!\!\!~\pub{ICLR21} & 72.41  & 69.14  & 65.50  & 57.60 & - & - & - & - \\
         AEL~\cite{hu2021semi}\!\!\!\!&\!\!\!~\pub{NeurIPS21} & - & - & - & - & 80.29 & 78.06 & 77.57 & 77.20 \\
         CPS~\cite{chen2021semi}\!\!\!\!&\!\!\!~\pub{CVPR21} & 75.88  & 71.71  & 67.42  & 64.07 & 78.64 & 77.55 & 75.83 & 72.18 \\
         PC2Seg~\cite{zhong2021pixel}\!\!\!\!&\!\!\!~\pub{ICCV21}  & 73.05  & 69.78  & 66.28  & 57.00 & - & - & - & - \\
         PS-MT~\cite{liu2022perturbed}\!\!\!\!&\!\!\!~\pub{CVPR22}  & 78.42  & 76.57  & 69.58  & 65.80 & 79.76 & 78.72 & 78.20 & 75.50 \\
         ST++~\cite{yang2021st++}\!\!\!\!&\!\!\!~\pub{CVPR22}  & 77.30  & 74.60  & 71.00  & 65.20 & - & 77.90 & 77.90 & 74.70 \\
         U2PL~\cite{wang2022semi}\!\!\!\!&\!\!\!~\pub{CVPR22}  & 76.16  & 73.66  & 69.15  & 67.98 & 80.50 & 79.30 & 79.01 & 77.21 \\
         GTA-Seg~\cite{jin2022semi}\!\!\!\!&\!\!\!~\pub{NeurIPS22}  & 78.37 & 75.57 & 73.16 & 70.02 & 81.01 & 80.57 & 80.47 & 77.82 \\
         MKD~\cite{yuan2022semi}\!\!\!\!&\!\!\!~\pub{NeurIPS22}  & 78.66 & 76.76 & 74.63 & 69.10 & 80.60 & 79.55 & 79.74 & 78.44 \\
         \hline
         \multicolumn{2}{|l||}{~~~~~~\textbf{\texttt{Ours}}} & \reshlb{79.39}{0.73} & \reshlb{77.93}{1.17} & \reshlb{76.66}{2.03} & \reshlb{73.25}{4.15} & \reshlb{81.00}{0.40} & \reshlb{80.62}{1.07} & \reshlb{80.24}{0.50} & \reshlb{79.65}{1.21} \\
\multicolumn{2}{|l||}{~~~~~~\textbf{\texttt{Ours}}~+~AEL~\cite{hu2021semi}} & \reshl{79.56}{0.17} & \reshl{78.16}{0.23} & \reshl{76.86}{0.20} & \reshl{73.65}{0.41} & \reshl{81.11}{0.11} & \reshl{80.78}{0.16} & \reshl{80.47}{0.23} & \reshl{79.77}{0.12} \\
\multicolumn{2}{|l||}{~~~~~~\textbf{\texttt{Ours}}~+~MKD~\cite{yuan2022semi}} & \reshl{80.06}{0.67} & \reshl{78.43}{0.50} & \reshl{77.18}{0.52} & \reshl{74.70}{1.45} & \reshl{81.21}{0.21} & \reshl{80.95}{0.33} & \reshl{80.53}{0.29} & \reshl{80.08}{0.43} \\
         \hline
      \end{tabular}
   }
   \captionsetup{font=small}
   \caption{\small\textbf{Quantitative results} (\S\ref{sec:exp:seg}) on PASCAL VOC 2012~\cite{everingham2015pascal} \texttt{val}. All methods are built upon DeepLabV3+~\cite{chen2018encoder}-ResNet101~\cite{he2016deep}.}
   \label{table:pascal}
   \vspace{-2pt}
\end{table*}

\begin{table}[t]
   \centering\small
   \resizebox{\columnwidth}{!}{
      \tablestyle{7pt}{1.05}
      \begin{tabular}{|rl||cccc|}
         \thickhline
         \rowcolor{mygray}
         & & 1/2 & 1/4 & 1/8 & 1/16 \\
         \rowcolor{mygray}
         \multicolumn{2}{|c||}{\multirow{-2}{*}{Method}} & (1488) & (744) & (372) & (186) \\

         \hline\hline
         MT~\cite{tarvainen2017mean}\!\!\!\!\!\!&\!\!\!\!~\pub{NeurIPS17}   &78.59&76.53&73.71&68.08\\
         CCT~\cite{ouali2020semi}\!\!\!\!\!\!&\!\!\!\!~\pub{CVPR20}  &78.29&76.35&74.48&69.64\\
         GCT~\cite{ke2020guided}\!\!\!\!\!\!&\!\!\!\!~\pub{ECCV20}   & 78.58&76.45&72.96&66.90\\
         CutMixSeg~\cite{french2020semi}\!\!\!\!\!\!&\!\!\!\!~\pub{BMVC20}   & 78.95&77.24&75.83&72.13\\
         AEL~\cite{hu2021semi}\!\!\!\!\!\!&\!\!\!\!~\pub{NeurIPS21} & 80.28&79.01&77.90&75.83\\
         CPS~\cite{chen2021semi}\!\!\!\!\!\!&\!\!\!\!~\pub{CVPR21} & 76.81&74.58&74.31&69.78\\
         S-Baseline~\cite{yuan2021simple}\!\!\!\!\!\!&\!\!\!\!~\pub{ICCV21} & 78.70&77.80&74.10&-\\
         U2PL~\cite{wang2022semi}\!\!\!\!\!\!&\!\!\!\!~\pub{CVPR22}  &79.05&76.47&74.37&70.30\\
         GTA-Seg~\cite{jin2022semi}\!\!\!\!\!\!&\!\!\!\!~\pub{NeurIPS22}  & 76.08 & 72.02 & 69.38 & 62.95 \\
         MKD~\cite{yuan2022semi}\!\!\!\!\!\!&\!\!\!\!~\pub{NeurIPS22}  & 80.74 & 78.28 & 75.98 & 75.31 \\
         \hline
         \multicolumn{2}{|r||}{\textbf{\texttt{Ours}}~~} & \textbf{80.95} & \textbf{80.21} & \textbf{78.90} & \textbf{76.83} \\
         \hline
      \end{tabular}
   }
   \captionsetup{font=small}
   \caption{\small\textbf{Quantitative results} (\S\ref{sec:exp:seg}) on Cityscapes~\cite{cordts2016cityscapes} \texttt{val}.}
   \label{table:cityscapes}
   \vspace{-2pt}
\end{table}

\noindent\textbf{Implementation Detail.}
In practice, computing the full semantics of the universal quantification $\forall$ is infeasible due to the large learning corpora, where we use batch-training as sampling based approximation~\cite{van2022analyzing}.
To facilitate efficient distributed training, we implement diagnostic reasoning steps using matrix multiplications. Besides, we employ point-wise supervision~\cite{kirillov2020pointrend} to further optimize efficiency while maintaining high performance standards.

\section{Experiment}\label{sec:exp}

\subsection{Experimental Setup}
\noindent\textbf{Datasets.} We evaluate {\Ours} on standard datasets:
\setlist{nosep}
\begin{itemize}[leftmargin=*]
   \setlength{\topsep}{0pt}
   \setlength{\itemsep}{0pt}
   \setlength{\parsep}{0pt}
   \setlength{\parskip}{0pt}
   \setlength{\partopsep}{0pt}
   \setlength{\itemindent}{0pt}

   \item \textbf{PASCAL $_{\!}$VOC $_{\!}$2012}\tcite{everingham2015pascal} is a famous semantic segmenta- tion dataset, consisting \app \cnum{4}k samples in \textit{original} dataset, that are split into \cnum{1464}/\cnum{1449}/\cnum{1456} images for \texttt{train}/\\\texttt{val}/\texttt{test}, respectively. It provides annotations for \cnum{21} categories including the background, which are grouped into \cnum{4} superclasses.
   Following the conventions~\cite{chen2021semi,wang2022semi}, \cnum{9118} coarsely labeled images in SBD~\cite{hariharan2011semantic} are adopted to complement \texttt{train} data, namely the \textit{augmented} set.

   \item \textbf{Cityscapes}~\cite{cordts2016cityscapes} has \cnum{5}k fine-annotated urban scene im- ages, with \cnum{2975}/\cnum{500}/\cnum{1524} for \texttt{train}/\texttt{val}/\texttt{test}, that defines $19$ semantic categories organized within \cnum{6} superclasses. Please refer to the supplementary for details.

   \item \textbf{COCO}~\cite{lin2014microsoft} features dense annotations for \cnum{80} object categories (\eg, animals, furniture, \etc), within diverse indoor and outdoor scenes. COCO stands out as the largest among the three benchmarks, comprising \cnum{118}k/\cnum{5}k images for \texttt{train}/\texttt{val}. The dataset officially provides a three-level semantic hierarchy, covering \cnum{2}/\cnum{12}/\cnum{80} concepts. 

\end{itemize}

\noindent\textbf{Partition Protocol.} We conduct evaluation under standard partition protocols~\cite{zhong2021pixel,wang2022semi}. For PASCAL VOC/Cityscapes, we sample 1/2, 1/4, 1/8, and 1/16 of the whole training set as labeled data. For COCO, we use smaller ratios, \ie, 1/32, 1/64, 1/128, 1/256, 1/512, considering the larger size of the dataset. We adopt the same sampled data to the state-of-the-arts~\cite{wang2022semi,zou2020pseudoseg} to enable meaningful comparisons.

\noindent\textbf{Base Network Architecture.} We take DeepLabV3+~\cite{chen2018encoder} as base segmentation architecture (\ie, $h(\cdot)$, \textit{cf.\!}~\S\ref{sec:method}), where ResNet101~\cite{he2016deep} and Xception65~\cite{chollet2017xception} pretrained on ImageNet~\cite{deng2009imagenet} are adopted as the backbone networks.

\noindent\textbf{Training.} We implement {\Ours} on MMSegmentation~\cite{mmseg2020} following standardized training settings and common weak/strong data augmentations~\cite{yuan2022semi,wang2022semi,yang2021st++} (\ie, $\mathcal{A}$/$\alpha$, \textit{cf.\!}~\S\ref{sec:method}). For PASCAL VOC/Cityscapes/COCO, images are cropped to $513\!\times\!513$/$769\!\times\!769$/$513\!\times\!513$ and models are trained for \cnum{40}k/\cnum{40}k/\cnum{40}k iterations with \cnum{16}/\cnum{16}/\cnum{16} batch size. The SGD optimizer is adopted with a weight decay of {5e-4}/{5e-4}/{5e-4}. The learning rate is set to $0.0025$/\cnum{0.01}/\cnum{0.01}, and is scheduled following the polynomial annealing policy~\cite{chen2017rethinking}, with a power of \cnum{0.9}/\cnum{0.9}/\cnum{0.9}.
We set $\lambda\!=\!5$ to balance the unsupervised loss (\textit{c.f.}~Eq.~\ref{eq:floss}), and opt to assign $q\!=\!5$ for logic quantifier approximation (\textit{c.f.}~Eq.~\ref{eq:quantifier}).

\noindent\textbf{Inference.} Following~\cite{zhong2021pixel,yuan2022semi}, for PASCAL VOC/COCO, we keep the aspect ratio of test images and rescale the short side to $512$. For Cityscapes, we use sliding window inference with $769_{\!}\times_{\!}769$ window size. All results are reported without any test-time augmentation for the sake of fairness.

\noindent\textbf{Evaluation Metric.} Following the conventions~\cite{wang2022semi,chen2021semi}, we use mean intersection-over-union (mIoU) for evaluation. In diagnostic experiments, we further report mIoU at each hierarchy level $l$ (denoted as mIoU$^l$) for thorough analysis.

\noindent\textbf{Reproducibility.}
Our models are implemented in PyTorch. All experiments are conducted on four Tesla A100 GPUs.

\begin{table}[t]
   \centering\small
   \resizebox{\columnwidth}{!}{
      \tablestyle{4.5pt}{1.05}
      \begin{tabular}{|rl||ccccc|}
         \thickhline
         \rowcolor{mygray}
         & & 1/32  &  1/64 & 1/128 & 1/256 & 1/512 \\
         \rowcolor{mygray}
         \multicolumn{2}{|c||}{\multirow{-2}{*}{Method}} & (3697) & (1849) & (925) & (463) & (232) \\
         \hline\hline
         Supervised\!\!\!\!& & 42.24 & 37.80 & 33.60 & 27.96 & 22.94 \\
         PseudoSeg~\cite{zou2020pseudoseg}\!\!\!\!&\!\!~\pub{ICLR21} & 43.64 & 41.75 & 39.11 & 37.11 & 29.78 \\
         PC2Seg~\cite{zhong2021pixel}\!\!\!\!&\!\!~\pub{ICCV21}  & 46.05 & 43.67 & 40.12 & 37.53 & 29.94 \\
         MKD~\cite{yuan2022semi}\!\!\!\!&\!\!~\pub{NeurIPS22} & 47.25 & 45.50 & 42.32 & 38.04 & 30.24 \\
         \hline
         \multicolumn{2}{|r||}{\textbf{\texttt{Ours}}~~} & \textbf{50.51} & \textbf{48.83} & \textbf{45.35} & \textbf{40.28} & \textbf{33.07} \\
         \hline
      \end{tabular}
   }
   \captionsetup{font=small}
   \caption{\small\textbf{Quantitative results} (\S\ref{sec:exp:seg}) on COCO~\cite{lin2014microsoft} \texttt{val}, based on DeepLabV3+~\cite{chen2018encoder}-Xception65~\cite{chollet2017xception} architecture.}
   \label{table:coco}
   \vspace{-2pt}
\end{table}

\begin{table*}[t]
	\subfloat[{Key Components}\label{tab:abl:components}]{
		\tablestyle{3pt}{1.05}
		\begin{tabular}{|ll||cc|c|}\thickhline
			\rowcolor{mygray}
			 & & & & Speed \\
			\rowcolor{mygray}
			\multicolumn{2}{|c||}{\multirow{-2}{*}{Method}}  & \multirow{-2}{*}{mIoU$^2$} & \multirow{-2}{*}{mIoU$^1$} & \textit{min}. \\\hline\hline
Baseline&                                                            & - & 68.02 & 85.3 \\\hline
~+ Hierarchical Prediction\!&\!\!\!(\S\ref{sec:meth:use})       & 84.34 & 69.15 & 86.5\color{gray}{\tiny{~+1.1\%}} \\
~+ Diagnosis Computation  \!&\!\!\!(\S\ref{sec:meth:define:DC})    & 86.31 & 71.83 & 88.4\color{gray}{\tiny{~+3.7\%}} \\
~+ Fuzzy Diagnosis Likelihood  \!&\!\!\!(\S\ref{sec:meth:measure})   & 87.91 & 73.25 & 89.3\color{gray}{\tiny{~+4.7\%}} \\
			\hline
			\multicolumn{2}{c}{}
		\end{tabular}
	}\hfill
	\subfloat[{Semantic Logic Rules}\label{tab:abl:constraints}]{
		\tablestyle{3pt}{1.05}
		\begin{tabular}{|ccc||cc|}\thickhline
			\rowcolor{mygray}
			$\Kcal_\text{C}$ & $\Kcal_\text{D}$ & $\Kcal_\text{E}$ & &  \\
			\rowcolor{mygray}
			Eq.~\ref{eq:rule:composition} & 	Eq.~\ref{eq:rule:decomposition} & 	Eq.~\ref{eq:rule:exclusion}  & \multirow{-2}{*}{mIoU$^2$} & \multirow{-2}{*}{mIoU$^1$} \\ \hline\hline
			& &                    	 & 84.34 & 69.15 \\
			\cmark & &               & 86.92 & 72.25 \\
			&\cmark &          	    & 87.10 & 72.47 \\
			& &\cmark           	    & 85.00 & 71.31 \\
			\cmark & \cmark&\cmark   & 87.91 & 73.25 \\
			\hline
		\end{tabular}
	}\hfill
	\subfloat[{Resolution Strategy}\label{tab:abl:strategy}]{%
		\tablestyle{3pt}{1.05}
		\begin{tabular}{|c||cc|}\thickhline
			\rowcolor{mygray}
			Resolution &  &  \\
			\rowcolor{mygray}
			Strategy & \multirow{-2}{*}{mIoU$^2$} & \multirow{-2}{*}{mIoU$^1$} \\ \hline\hline
			Uniform & 86.31 & 71.83 \\
         Predictive & 86.65 & 72.05 \\
			Greedy & 87.29 & 72.61 \\
			Sampling & 87.91 & 73.25 \\
			\hline
			\multicolumn{2}{c}{}
		\end{tabular}
	}\hfill
	\vspace{-1pt}
	\captionsetup{font=small}
   \caption{\small \textbf{Ablative experiments} on PASCAL VOC 2012~\cite{everingham2015pascal} \texttt{val} with 1/16 \textit{augmented} training set. Please refer to \S\ref{sec:exp:dig} for more details.}
	\label{tab:ablations}
	\vspace{-2pt}
\end{table*}

\subsection{Quantitative Comparison Result}\label{sec:exp:seg}

\noindent\textbf{PASCAL VOC 2012 \textit{original}.} Table~\ref{table:pascal} (left) summarizes the quantitative comparisons under varying label amounts, from which we take three major observations:
\textbf{First}, Our method indeed surpasses the SOTAs across all settings and establishes the new state-of-the-arts of \bpo{79.39}/\bpo{77.93}/ \bpo{76.66}/\bpo{73.25} under 1/2-1/16 partitions, indicating the effectiveness of our neural-logic framework.
\textbf{Second}, The superiority of our method is more significant on fewer labeled data, with the largest margin achieved when only 1/16 labels are available, \ie, \bpo{4.15} over MKD, demonstrating its potential in extremely label-scarce scenarios.
\textbf{Third}, Our framework is fully compatible with mainstream SSL methods~\cite{yuan2022semi,wang2022semi}. The performance can be consistently lifted when equipped with additional perturbation techniques.

\noindent\textbf{PASCAL VOC 2012 \textit{augmented}.} Table~\ref{table:pascal} (right) demonstrates the comparison results on the PASCAL VOC 2012 \texttt{val} using \textit{augmented} set for training, where our method again sets new state-of-the-arts across all partition protocols, yielding an average gain of \bpo{0.80} upon the previous SOTA~\cite{yuan2022semi}.
When only provided with scarce labeled data, \ie, \cnum{92} labeled images, our method still performs impressive owing to the compactly incorporated symbolic logic.

\noindent\textbf{Cityscapes.} Table~\ref{table:cityscapes} quantitatively compares our method against the competitors on Cityscapes \texttt{val}. In spite of the presence of complex street scenes, our method still delivers a solid overtaking trend across different partitions, with an averaged advancement of \bpo{1.65} in terms of mIoU.

\noindent\textbf{COCO.} Table~\ref{table:coco} presents the model performance on COCO \texttt{val}. As observed, by incorporating the large semantic hierarchy in COCO, {\Ours} provides even greater performance gains against the leading method (\ie, MKD~\cite{yuan2022semi}) across all partitions by \bpo{2.94} mIoU on average. The experimental results confirm again the efficacy of {\Ours}.

\vspace{3pt}
\subsection{Diagnostic Experiment}\label{sec:exp:dig}

For in-depth analysis, we perform a set of ablative studies on PASCAL VOC 2012~\cite{everingham2015pascal} \texttt{val} with 1/16 \textit{augmented} set.
Please refer to the supplementary for more experiments.

\noindent\textbf{Key Component Analysis.}
In Table~\ref{tab:abl:components}, we first validate the importance of our proposed components by attaching them one at a time.
The 1$^\textit{st}$ row reports the result of a bare baseline model - DeepLabV3+ with plain consistency regularization~\cite{sohn2020fixmatch}.
Next, in the 2$^\textit{nd}$ row, we convert the prediction mode from flat to hierarchical, which already boosts performance and supports our claim that hierarchical semantics can $_{\!}$provide $_{\!}$additional $_{\!}$training $_{\!}$signals $_{\!}$implicitly. $_{\!}$
Moreover, the 3$^\textit{rd}$ row gives the score when the minimal diagnosis set is further computed, and we uniformly sample one conflict from it to resolve. As seen, this leads to moderate improvement caused by the explicit introduction of symbolic knowledge that potentially resolves conflicts.
Finally, as shown in the 4$^\textit{th}$ row, through sampling from the diagnosis likelihood, the biggest improvement is achieved, demonstrating the necessity of the fuzzy measurement that guides the resolution.

\noindent\textbf{Semantic Logic Rules.}
Then, we investigate the effectiveness of logic-induced hierarchy rules (\S\ref{sec:meth:use}) in Table~\ref{tab:abl:constraints}. Starting from the baseline (1$^\textit{st}$ row), we individually add Composition (\textit{cf.\!}~Eq.~\ref{eq:rule:composition}), Decomposition (\textit{cf.\!}~Eq.~\ref{eq:rule:decomposition}), and Exclusion (\textit{cf.\!}~Eq.~\ref{eq:rule:exclusion}) rules, denoted as $\Kcal_\text{C},\Kcal_\text{D},\Kcal_\text{E}$, \textit{resp.}, into the proposed framework, resulting in the scores listed in the 2$^\textit{nd}$ to 4$^\textit{th}$ rows. The last row exhibits the outcome achieved with our full training regime.
Upon examining the table, three observations can be made.
\textbf{First}, incorporating each of the logic rule results in consistent performance gains, indicating that the set of rules captures diverse facets of visual semantics, and indeed benefits SSL models within our symbolic resolution framework.
\textbf{Second}, the best performance is attained by combining all three logic rules, highlighting the significance of comprehensive interpretation of structured semantic concepts.
\textbf{Third}, this also implies that integrating extra symbolic knowledge has great potential to further enhance current sub-symbolic SSL.

\noindent\textbf{Resolution Strategy.}
Table~\ref{tab:abl:strategy} reveals the impact of conflict resolution strategies (\S\ref{sec:meth:measure}).
The default strategy, `Sampling', utilizes the Monte Carlo method to estimate the actual posterior distribution $P(\omega|\Scal)$ (\textit{cf.\!}~Eq.~\ref{eq:likelihood}).
Here we investigate three alternatives: `Uniform', which assigns equal weight to all valid diagnoses;
`Predictive', which samples solely based on predictive probability $P(o|\Scal)$ without considering the degree of conflicts;
`Greedy', which resolves conflicts by prioritizing the highest probability according to $P(\omega|\Scal)$.
The results show that `Predictive', `Greedy', and `Sampling' all outperform `Uniform', indicating the necessity of likelihood modeling. Of these strategies, `Sampling' stands out as the most effective, providing strong evidence for its capacity to escape from the spurious corrections.

\noindent\textbf{Training Speed.} For completeness, we include the training time of 10K iterations in the last column of Table~\ref{tab:abl:components}.~Our training $_{\!}$regime $_{\!}$(the $_{\!}$last $_{\!}$row) $_{\!}$incurs $_{\!}$a $_{\!}$trivial $_{\!}$delay $_{\!}$of \app 4.7\%.

\noindent\textbf{Inference $_{\!}$Speed.} $_{\!}$During $_{\!}$inference, $_{\!}$the $_{\!}$ancestral $_{\!}$classes $_{\!}$in $_{\!}$the $_{\!}$hierarchical $_{\!}$classification $_{\!}$head $_{\!}$can $_{\!}$be $_{\!}$safely $_{\!}$disregarded without introducing additional computational burden.

\vfill
\section{Conclusion}
This paper introduces {\Ours}, a neural-logic SSL framework that consolidates the benefits from both symbolic reasoning and sub-symbolic learning. Through resolving conflicts within pseudo labels using logic-induced diagnoses, {\Ours} systematically compiles rich symbolic knowledge into the neural training pipeline.
Experimental findings illustrate that {\Ours} outperforms existing SSL frameworks, especially in label scarce settings. The enhanced flexibility and generality of {\Ours} showcase the immense potential of this holistic neural-logic paradigm in pixel-wise semi-supervised learning.
We believe this paper opens a new avenue for future exploration in the field.

\clearpage
{\small
\bibliographystyle{unsrt}
\bibliography{reference}
}

\clearpage
\appendix

\setlist{nosep}  
\begin{itemize}[leftmargin=*]
	\setlength{\itemsep}{0pt}
	\setlength{\parsep}{-2pt}
	\setlength{\parskip}{-0pt}
	\setlength{\leftmargin}{-10pt}
  \item \S\ref{sec:fol}: Detailed introduction of First-order Logic.
  \item \S\ref{sec:algorithm}: Detailed {\Ours} algorithm.
  \item \S\ref{sec:hier}: Detailed label hierarchy.
  \item \S\ref{sec:exp}: More experimental results.
  \item \S\ref{sec:qualitative}: More qualitative visualization.
\end{itemize}

\section{Detailed Introduction of First-order Logic}\label{sec:fol}

First-order logic (FOL), also known as predicate logic, is a formal language system used for representing and reasoning about statements involving objects and their properties. FOL extends propositional logic by introducing predicates, which are functions that take one or more arguments and return a truth value.
The syntax of FOL is defined by a set of symbols, including variables, constants, function symbols, predicate symbols, and logical connectives. The basic components of FOL are (please also refer to \S\textcolor{red}{3.2.1}):

\noindent$\bullet$ \textit{Constants:} Objects that are always present and have a fixed interpretation. For example, a specific pixel $x_i$.

\noindent$\bullet$ \textit{Variables:} Symbols that represent objects whose identity is not specified. For example, pixel data sample $x$.

\noindent$\bullet$ \textit{Quantifiers:} Symbols that specify the extent of a statement's applicability, either `for all' ($\forall$) or `there exists' ($\exists$).

\noindent$\bullet$ \textit{Connectives:} Symbols that combine statements to form more complex ones, \eg, negation ($\neg$), conjunction ($\land$), disjunction ($\lor$), implication ($\rightarrow$), and biconditional ($\leftrightarrow$).

\noindent$\bullet$ \textit{Predicates:} Expressions that assert a relationship between objects. For example, $Cat(x)$: $x$ is a Cat.

In FOL, a formula is formed by combining atomic formulas using logical connectives and quantifiers. An atomic formula is typically a predicate applied to a set of terms, and takes the form $P(t_1,\dots,t_n)$, where $P$ is a predicate symbol of arity $n$ and $t_1,\dots,t_n$ are terms, which may be variables, constants, or function symbols applied to terms.
The semantics of FOL are determined by a truth function that assigns a truth value to each formula based on the values of its subformulas and the domain of discourse. As an example, in our main paper, we define a unary predicate $t(\cdot)$ that takes a single input $o$ and evaluates whether $o$ is logically consistent with the symbolic knowledge. The truth value of a formula of the form $\forall o,t(o)$ is true if and only if $t(o)$ is true for all objects in the domain of discourse.

\begin{algorithm}[t]
  \caption{{\Ours} Pseudocode, PyTorch-like style}
  \label{alg:code}
  \definecolor{codeblue}{rgb}{0.25,0.5,0.5}
  \definecolor{codekw}{rgb}{0.85, 0.18, 0.50}
  \lstset{
     backgroundcolor=\color{white},
     basicstyle=\fontsize{7.5pt}{7.5pt}\ttfamily\selectfont,
     columns=fullflexible,
     breaklines=true,
     captionpos=b,
     commentstyle=\fontsize{7.5pt}{7.5pt}\color{codeblue},
     keywordstyle=\fontsize{7.5pt}{7.5pt}\color{codekw},
  }
  \begin{lstlisting}[language=python]
  # H: neural network predictor
  # K: knowledge base of first-order logic rules
  # L: loss function, e.g., binary cross entropy loss
  # m: trade-off hyperparameter
  
  # load minibatch from labeled and unlabeled dataset
  # s_x, u_x: labeled and unlabeled datapoints
  # o_s: ground-truth labels
  for (s_x, o_s), u_x in loader:
     # Pose assumptions on observations
     p_s_o = H(s_x).sigmoid() # supervised prediction
     p_u_o = H(u_x).sigmoid() # unsupervised prediction
  
     loss_s = L(p_s_o, o_s) # supervised loss
  
     p_o = CONCAT([p_s_o, p_u_o])
     o = p_o.binarize() # original pseudo label
  
     # Compute diagnoses (Eq.4-5)
     Omega = DIAGNOSE(o, K)
  
     # Calculate diagnosis likelihood (Eq.7,8,11)
     p_Omega = FUZZY_LIKELIHOOD(Omega, p_o, K)
     omega = SAMPLING(p_Omega) # sample from likelihood
  
     # Resolve conflicts within pseudo labels (Eq.6)
     o_r = RESOLUTION(o, omega)
  
     # Calculate unsupervised loss on revised labels
     loss_u = L(p_o, o_r)
  
     # Calculate overall training objective (Eq.3)
     loss = loss_s + m*loss_u
  
     # Update the sub-symbolic network parameters
     loss.backward()
     update(H.params)
  
  \end{lstlisting}
\end{algorithm}

\section{Detailed {\Ours} Algorithm}\label{sec:algorithm}
Algorithm~\ref{alg:code} provides the pseudocode for {\Ours}. By integrating symbolic reasoning into the neural learning process, {\Ours} enhances the model's ability to align with background knowledge, correct conflicting predictions, and improves its predictive accuracy in SSL tasks.

\section{Detailed Label Hierarchy}\label{sec:hier}

\begin{figure}[t]
  \begin{center}
      \includegraphics[width=\linewidth]{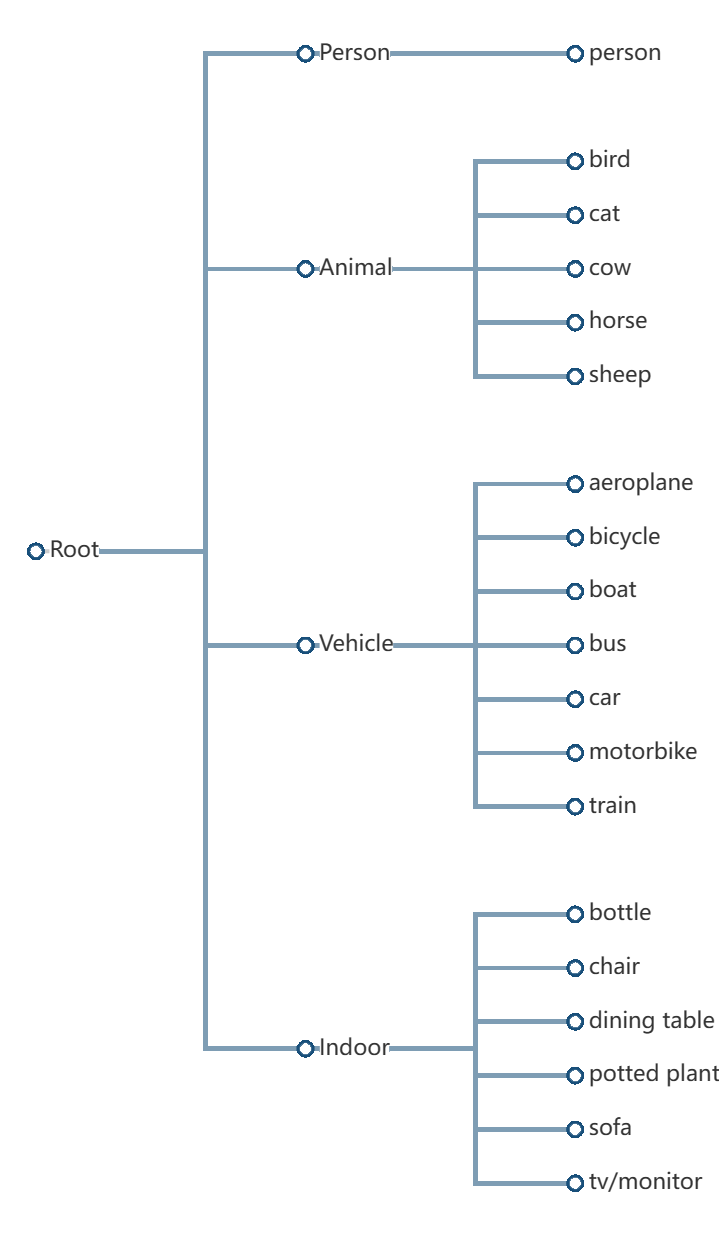}
  \end{center}
  \vspace{7.2ex}
  \captionsetup{font=small}
  \caption{\small \textbf{Official label hierarchy} of PASCAL VOC 2012~\cite{everingham2015pascal}. Please check more details at \url{http://host.robots.ox.ac.uk/pascal/VOC/}.}
  \label{fig:hier:pascal}
\end{figure}

\begin{figure}[t]
  \begin{center}
      \includegraphics[width=\linewidth]{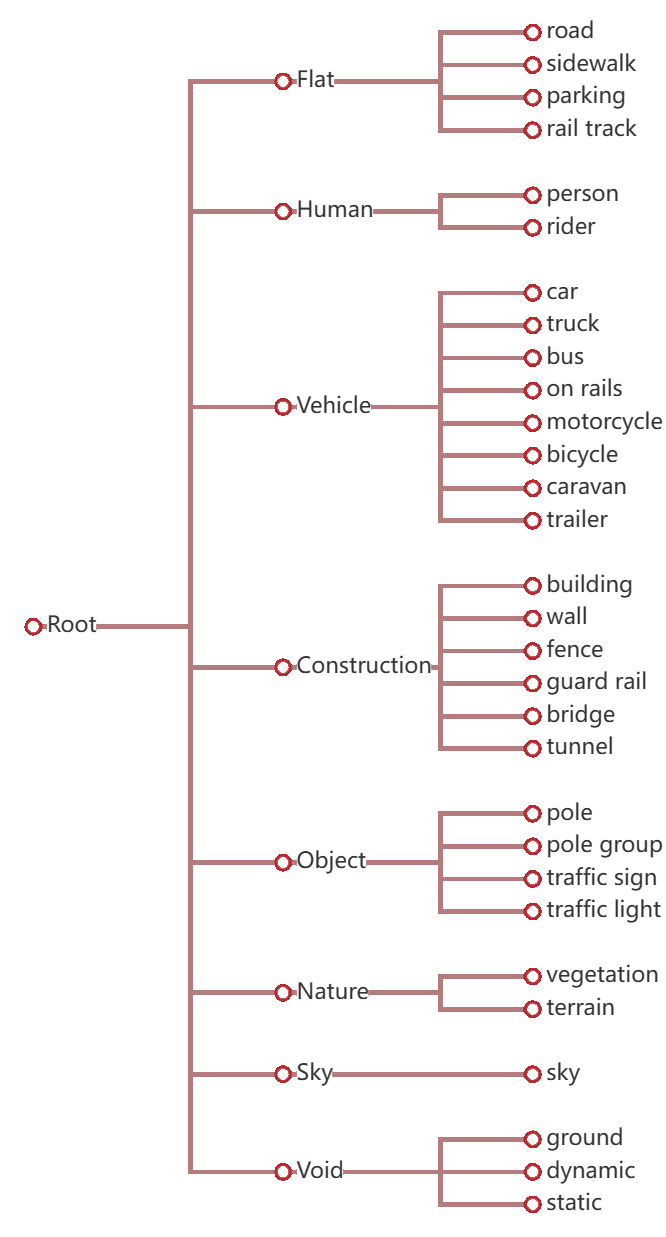}
  \end{center}
  \captionsetup{font=small}
  \caption{\small \textbf{Official label hierarchy} of Cityscapes~\cite{cordts2016cityscapes}. Please find more details at \url{https://www.cityscapes-dataset.com/dataset-overview/}.}
  \label{fig:hier:cityscapes}
\end{figure}

\begin{figure*}[t]
  \begin{center}
      \includegraphics[width=.75\linewidth]{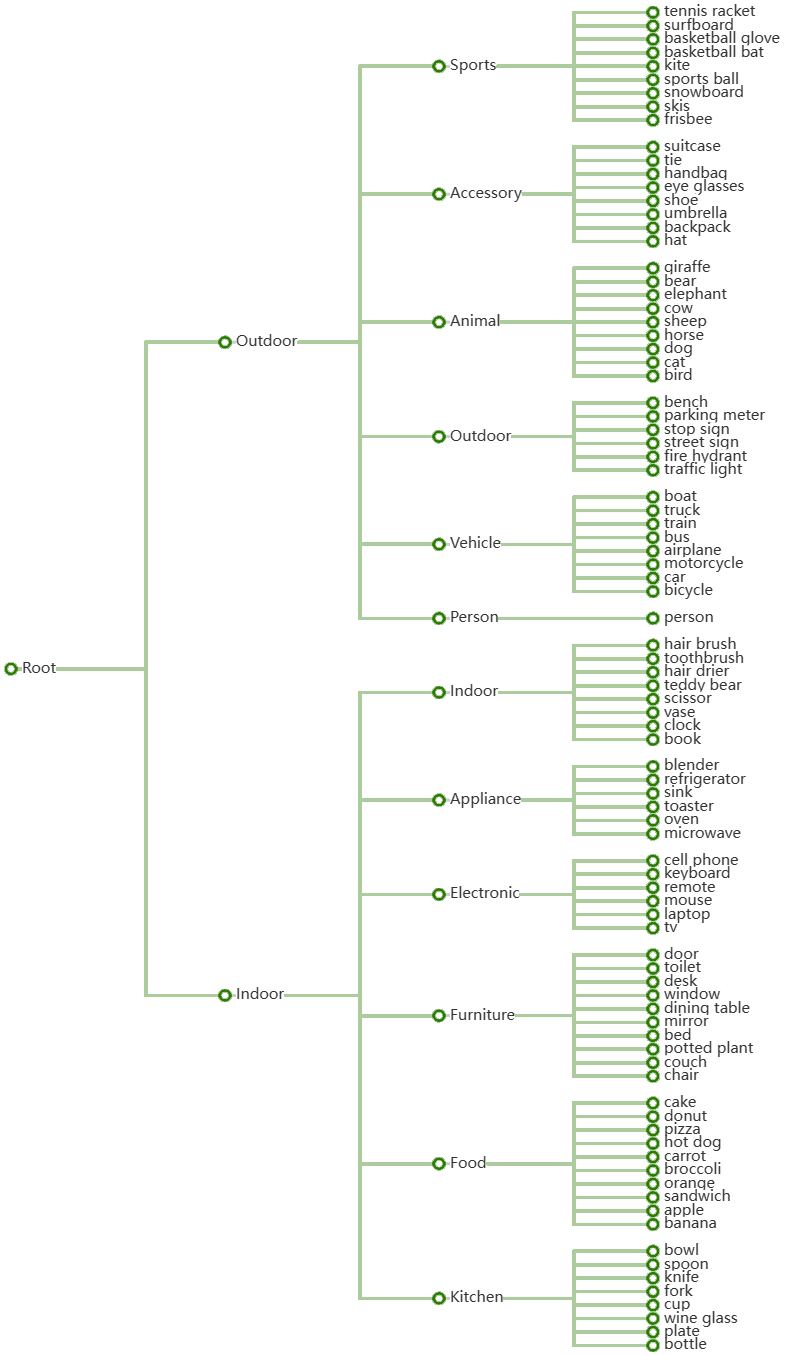}
  \end{center}
  \captionsetup{font=small}
  \caption{\small \textbf{Official label hierarchy} of COCO~\cite{lin2014microsoft}. Please find more details at \url{https://cocodataset.org/}.}
  \label{fig:hier:coco}
\end{figure*}

In this paper, we leverage the label hierarchy present in each dataset to derive the logic rules. 
To this end, we leverage the official structured label hierarchies for the PASCAL VOC 2012\footnote{\url{http://host.robots.ox.ac.uk/pascal/VOC/}}~\cite{everingham2015pascal}, Cityscapes\footnote{\url{https://www.cityscapes-dataset.com/dataset-overview/}}~\cite{cordts2016cityscapes}, and COCO\footnote{\url{https://github.com/nightrome/cocostuff}}~\cite{lin2014microsoft} datasets.
To represent the most general concept, we introduce a virtual root node labeled \texttt{Root}. The detailed hierarchies for PASCAL VOC 2012, Cityscapes, and COCO are illustrated in Fig.~\ref{fig:hier:pascal}, Fig.~\ref{fig:hier:cityscapes}, and Fig.~\ref{fig:hier:coco}, respectively.

\section{More Experimental Results}\label{sec:exp}

\begin{figure}[t]
   \definecolor{sheep}{RGB}{192,128,0}
   \begin{center}
      \includegraphics[width=1\linewidth]{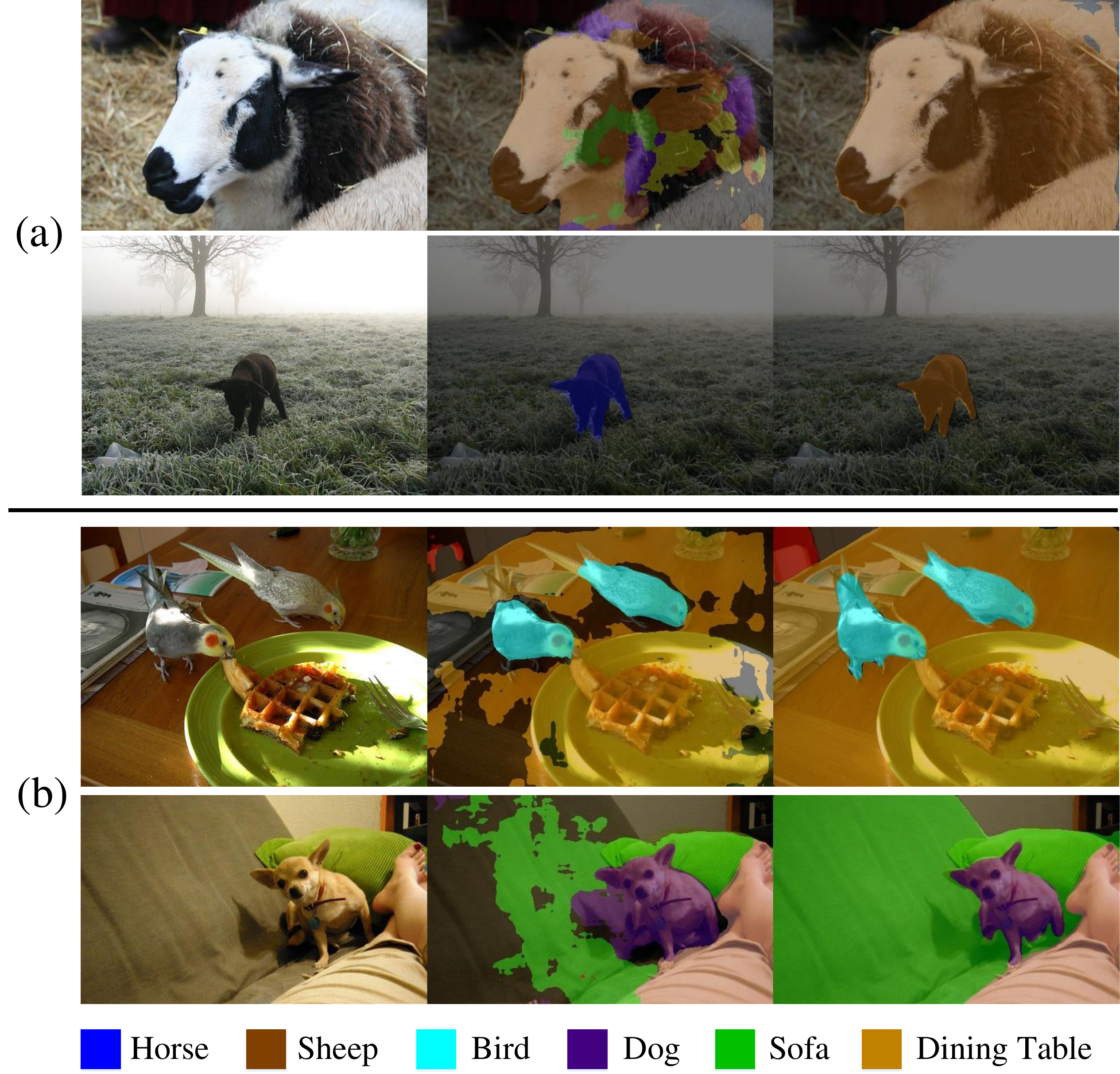}
  \end{center}
  
  \vspace{-10pt}
  \captionsetup{font=small}
  \caption{\small \textbf{Pseudo labels} obtained with confidence thresholding (middle) and {\Ours} (right). There are two key limitations associated with confidence thresholding. 
  (a) Confidence thresholding is not able to rectify potentially incorrect predictions, leading to the generation of inaccurate pseudo-labels. For example, as shown in the second column, the error accumulation ultimately leads to the misclassification of \textcolor{sheep}{sheep} as a \textcolor{blue}{cow}; 
  (b) The typically high threshold used in confidence thresholding results in a significant reduction in the number of generated pseudo-labels.}
  \label{fig:quality}
\end{figure}

\noindent\textbf{Semantics of Label Hierarchy.} 
We further examine the impact of hierarchical structure, which is used in deriving the Composition, Decomposition, and Exclusion  rules of visual concepts. By default, we use the official label hierarchies defined in PASCAL VOC 2012, Cityscapes, and COCO  (\textit{c.f.\!}~\S\ref{sec:hier}). We additionally explore an alternative where a random hierarchy was constructed with the same number of superclasses and classes per superclass as the official hierarchy. The results are presented in the table below.
For completeness, we also include the baseline without employing our framework. 
Our findings indicate that when using the randomly constructed hierarchy, the achieved results are only comparable or even slightly worse than the baseline, due to the presence of noisy supervision. This result again shows that the structured semantic concepts and derived set of logic rules are indeed helpful in the semi-supervised learning of semantic segmentation models.

\vspace{-4pt}
\begin{table}[h]
  \centering\small
  \resizebox{0.63\columnwidth}{!}{
  \tablestyle{4pt}{1.05}
		\begin{tabular}{|c||cc|}\thickhline
			\rowcolor{mygray}
			Label Hierarchy & \multirow{-1}{*}{mIoU$^2$} & \multirow{-1}{*}{mIoU$^1$} \\ \hline\hline
			None & - & 68.02 \\
			Random & 76.13 & 67.11 \\
			Official & 87.91 & 73.25 \\
			\hline
			\multicolumn{2}{c}{}
		\end{tabular}
    }
    \vspace{-6pt}
    \captionsetup{font=small}
    \caption{\small \textbf{Impact of semantics within label hierarchy}, evaluated on PASCAL VOC 2012~\cite{everingham2015pascal} \texttt{val} with 1/16 \textit{augmented} set.}
    \label{table:cityscapes}
\end{table}

\section{More Qualitative Visualization}\label{sec:qualitative}
\noindent\textbf{Pseudo-label Quality.} 
In Fig.~\ref{fig:quality}, we present the visualization that compares the pseudo labels generated by our {\Ours} with those obtained through the confidence thresholding method.
Our visualization highlights two key advantages of {\Ours}. 
First, as demonstrated in Fig.\,\ref{fig:quality}\,(a), the integration of conflict resolution mechanism enables {\Ours} to rectify possibly erroneous predictions, leading to more precise and cohesive pseudo-labels that align with the existing knowledge.
Second, Fig.\,\ref{fig:quality}\,(b) illustrates that the confidence thresholding method generates significantly fewer pseudo-labels compared to the {\Ours} framework. This limitation, in turn, hinders the effectiveness of semi-supervised learning process, which relies on large amounts of high-quality pseudo-labeled data.

\begin{figure}[t]
  \begin{center}
      \includegraphics[width=1\linewidth]{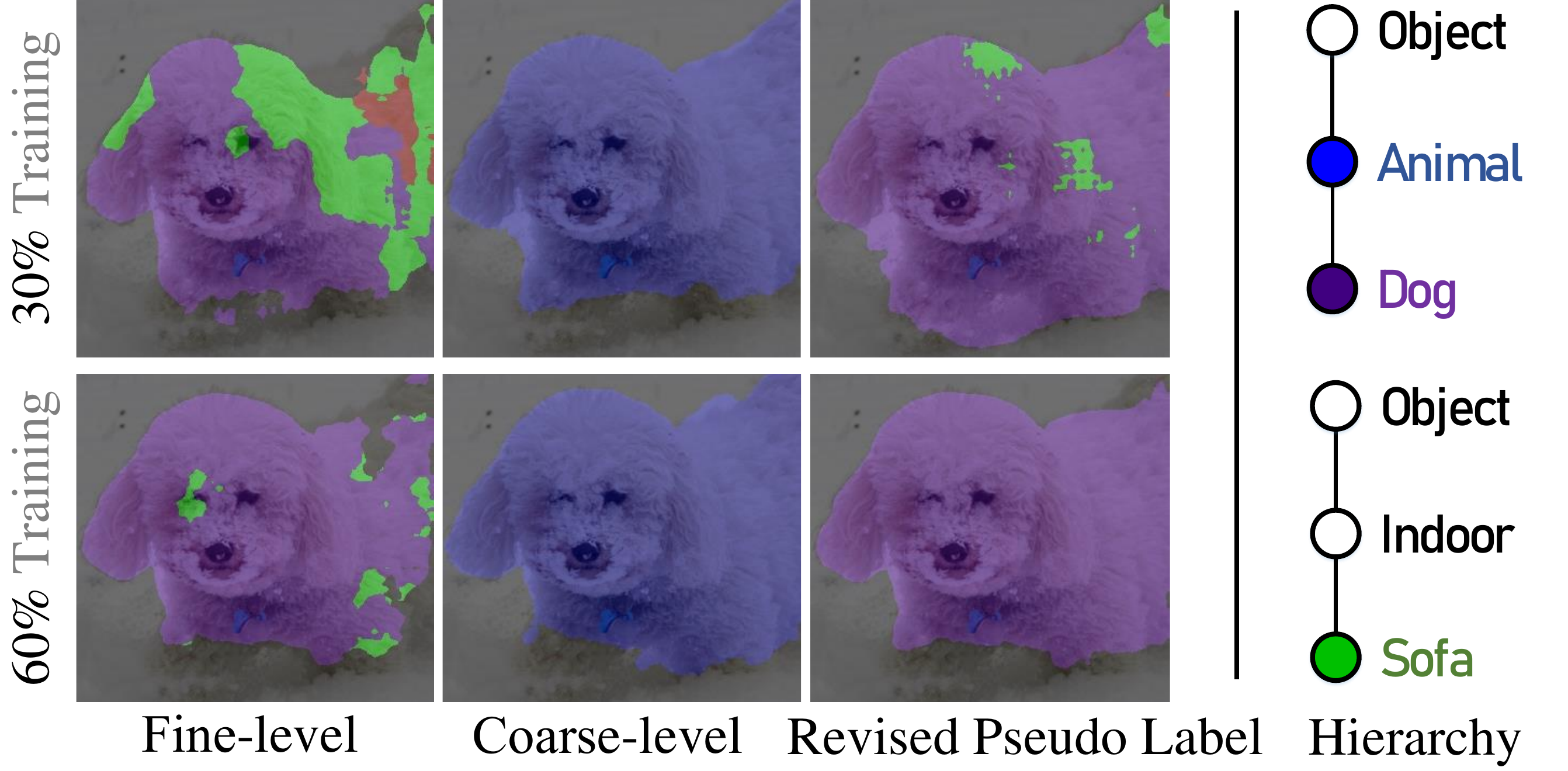}
  \end{center}
  \vspace{-10pt}
  \captionsetup{font=small}
  \caption{\small \textbf{Pseudo labels} \textit{before} and \textit{after} logic-induced diagnostic reasoning across training iterations.}
  \label{fig:int}
\end{figure}

\noindent\textbf{Diagnostic Reasoning.}
To further depict the effect of logic-induced diagnostic reasoning, we illustrate the process \wrt training iterations on one typical example in Fig.~\ref{fig:int}. 
Initially (30\% training), a substantial portion of the pseudo labels undergo revisions based on label hierarchy, resulting in improved accuracy. As training progresses (60\% training), the revised pseudo labels become more accurate, requiring fewer modifications. This demonstrates {\Ours}'s efficacy in refining and enhancing pseudo labels iteratively.

\begin{figure*}[t]
  \begin{center}
    \includegraphics[width=1 \linewidth]{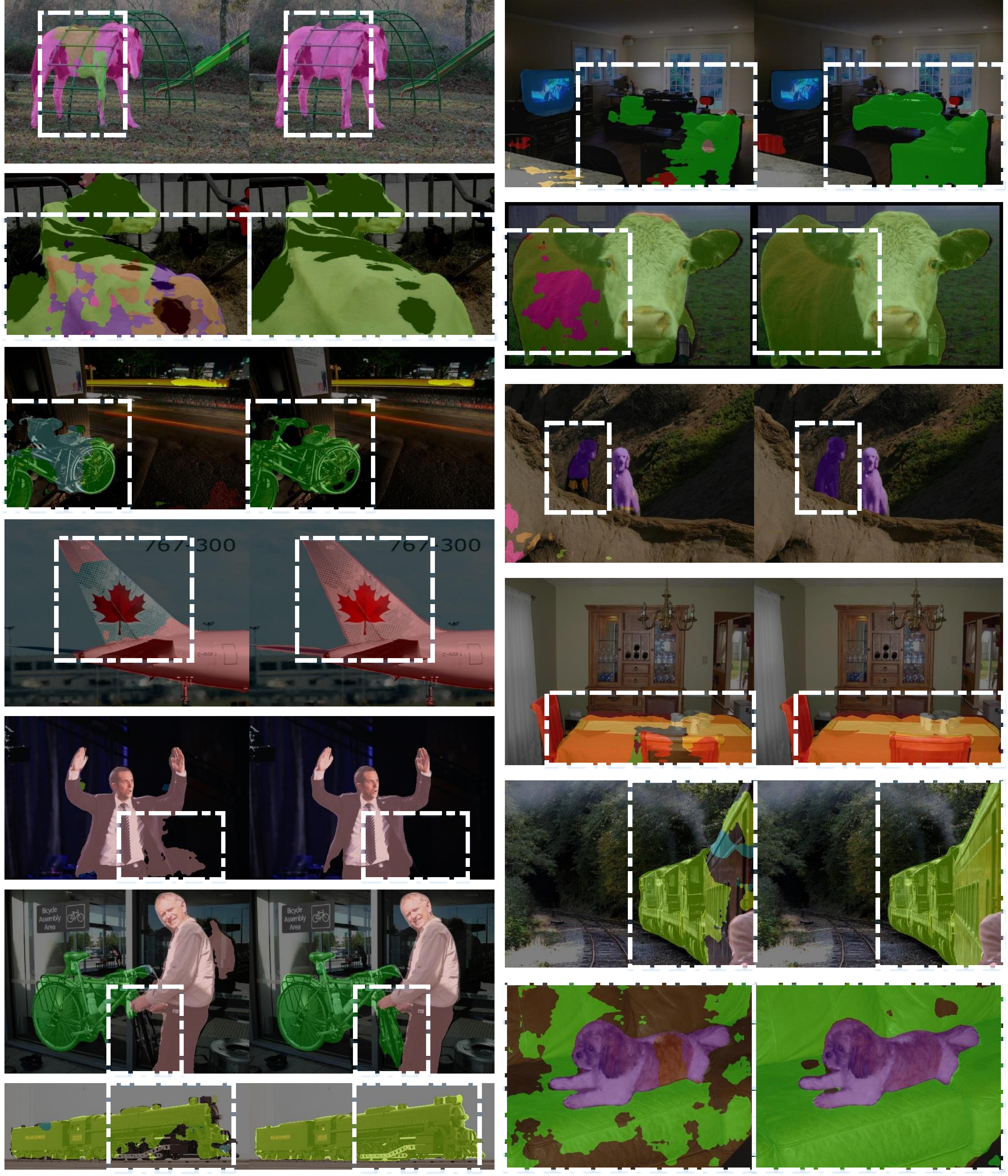}
  \end{center}
\caption{\textbf{Qualitative results} (\S\ref{sec:qualitative}) obtained from confidence thresholding (left) and {\Ours} (right) methods with DeepLabV3+~\cite{chen2018encoder} as the basic segmentation architecture on PASCAL VOC 2012~\cite{everingham2015pascal}.}
\label{fig:pascal}
\end{figure*}

\begin{figure*}[t]
  \begin{center}
    \includegraphics[width=1 \linewidth]{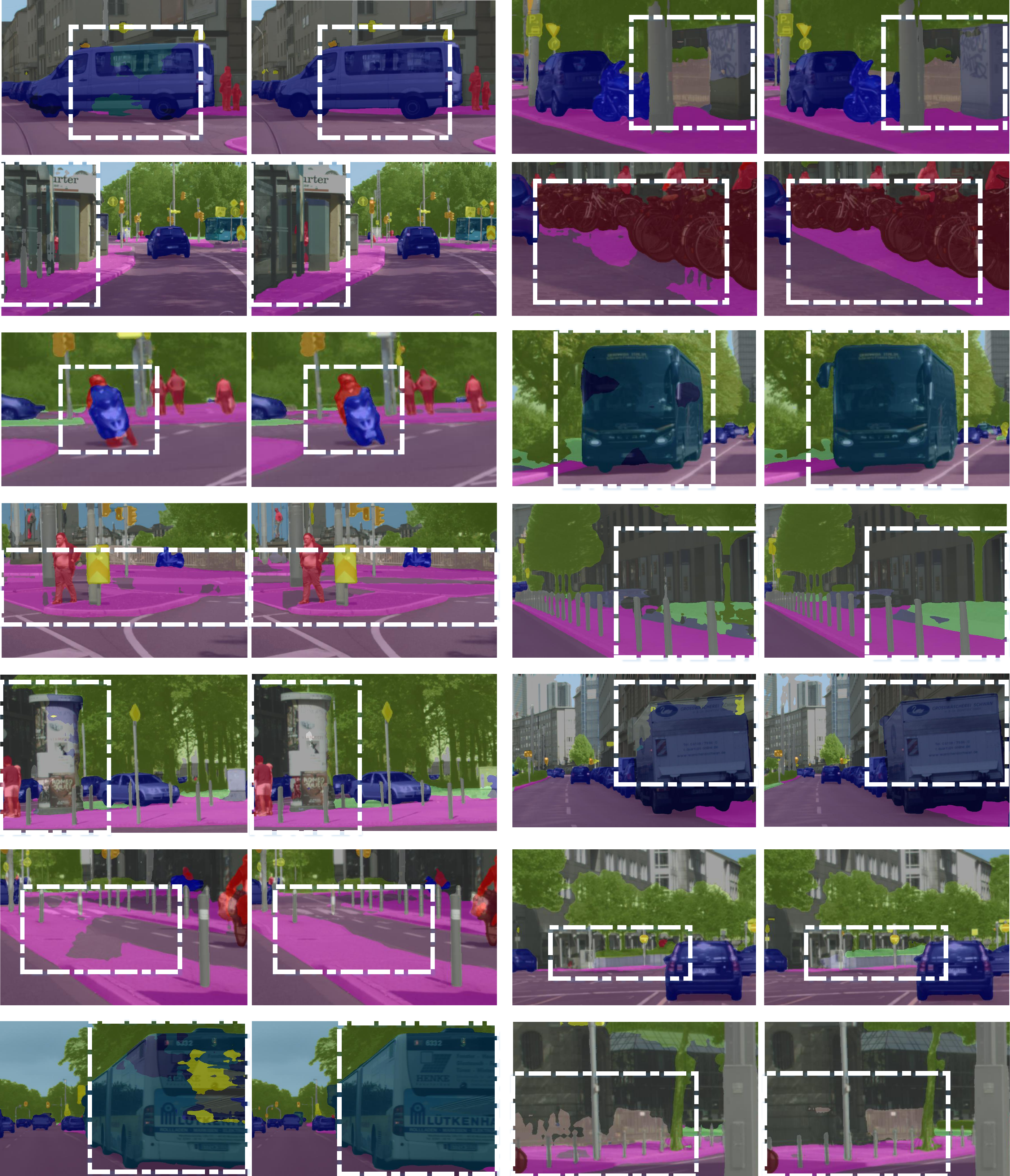}
  \end{center}
\caption{\textbf{Qualitative results} (\S\ref{sec:qualitative}) obtained from confidence thresholding (left) and {\Ours} (right) methods with DeepLabV3+~\cite{chen2018encoder} as the basic segmentation architecture on Cityscapes~\cite{cordts2016cityscapes}.}
\label{fig:cityscapes}
\end{figure*}

\noindent\textbf{Qualitative Results.} We illustrate representative qualitative results of our method and confidence thresholding upon the DeepLabV3+~\cite{chen2018encoder}, on the PASCAL VOC~\cite{everingham2015pascal} (Fig.~\ref{fig:pascal}), and Cityscapes~\cite{cordts2016cityscapes} (Fig.~\ref{fig:cityscapes}). 
It is evident that {\Ours} produces more accurate predictions attributed to the successful incorporation of symbolic visual semantics, which helps to resolve ambiguous classes and subtle textures that often cause confusion for the baselines.

\end{document}